\documentclass{article} %
\usepackage{lot-preprint,times}

\usepackage{amsmath,amsfonts,bm}

\def\Secref#1{Section~\ref{#1}}

\def\eqref#1{equation~\ref{#1}}

\def\1{\bm{1}}

\DeclareMathAlphabet{\mathsfit}{\encodingdefault}{\sfdefault}{m}{sl}
\SetMathAlphabet{\mathsfit}{bold}{\encodingdefault}{\sfdefault}{bx}{n}

\usepackage{graphicx}
\usepackage{hyperref}   
\usepackage{url}
\usepackage{booktabs}
\usepackage{multirow}
\usepackage{wrapfig}
\usepackage{needspace}
\usepackage{placeins}
\usepackage{etoc}

\let\cite\citep

\title{Level-of-token Diffusion}

\author{%
\textbf{Kiyohiro Nakayama}$^{1}$ \quad
\textbf{Brian Chao}$^{1}$ \quad
\textbf{Jan Ackermann}$^{1}$ \quad
\textbf{Hansheng Chen}$^{1}$ \\
\textbf{Federico Tombari}$^{2}$ \quad
\textbf{Leonidas Guibas}$^{1}$ \quad
\textbf{Lior Yariv}$^{1,*}$ \quad
\textbf{Gordon Wetzstein}$^{1}$\thanks{Equal advising.} \\[0.5em]
$^{1}$Stanford University \qquad $^{2}$Google
}

\hypersetup{
  hidelinks,
  pdftitle={Level-of-Token Diffusion},
  pdfauthor={Kiyohiro Nakayama, Brian Chao, Jan Ackermann, Hansheng Chen, Federico Tombari, Leonidas Guibas, Lior Yariv, Gordon Wetzstein},
  pdfsubject={Multiresolution token layouts for efficient image and video generation},
  pdfkeywords={diffusion models, adaptive tokenization, image generation, video generation}
}
\begin{document}

\maketitle
\begin{figure}[th!]
    \centering
    \vspace{-24pt}
    \includegraphics[width=0.88\textwidth]{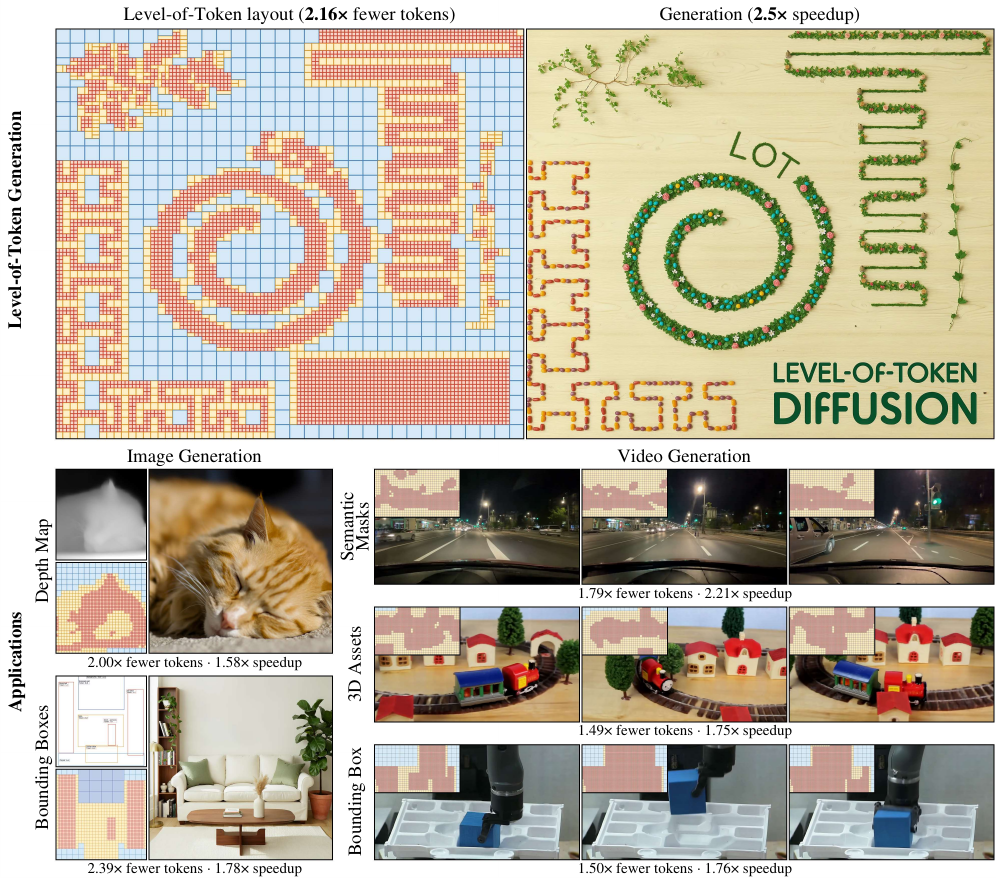}\vspace{-8pt}
    \caption{\textbf{Level-of-Token (LoT) Diffusion}
    allocates compute using a spatial token layout that specifies the spatially varying level of detail for efficient image and video generation.
    Derived from diverse spatial priors, token layouts guide detail allocation with speedups tied to the token budget.}
    \label{fig:teaser}
\end{figure}
\begin{abstract}\vspace{-6pt}
Image and video diffusion models allocate equal computation to every region, even when the intended scene calls for varying levels of detail. The spatial distribution of detail can often be anticipated before generation, indicating where computation can be reduced. We introduce \textit{Level-of-Token (LoT) Diffusion}, a framework that turns this knowledge into an explicit multiresolution token layout (\textit{Level-of-Token layout}) for adaptive and efficient generation. Tokens represent rectangular patches of varying sizes and shapes, allocating finer tokens where detail is needed and coarser tokens elsewhere. We adapt pretrained diffusion transformers to LoT layouts through a patch-wise asymmetric flow parametrization and embeddings for multiresolution tokens, preserving full-resolution flow prediction at every denoising step while processing only a reduced token sequence.
LoT Diffusion enables layout-adaptive generation while preserving pretrained generative priors. We demonstrate LoT with layouts derived from semantic masks, bounding boxes, texture variance, and depth-of-field cues, as well as agentic plans. Across image and video generation, LoT offers favorable quality–efficiency tradeoffs, with significant speedups determined by the layout’s token budget. Our project website is at \url{https://georgenakayama.github.io/lotdiffusion/}. 
\end{abstract}

\section{Introduction}
Image and video generative models produce remarkably high-quality content, yet generating high-resolution images and long videos remains computationally expensive. Much of this cost comes from diffusion transformers~\citep{peebles2023dit} processing long sequences of spatial tokens through attention and linear layers at every denoising step. 
Reducing this cost begins with recognizing that not all regions require the same level of detail: intricate textures, faces, and text require fine detail, while clear skies, blurred backgrounds, and flat surfaces can be synthesized more coarsely. 
Pretrained models nevertheless allocate equal computation to all regions.
Allocating resources according to the detail needed is well established in level-of-detail rendering~\citep{clark1976lod}, variable-rate shading~\citep{vaidyanathan2014coarse}, and image compression~\citep{shapiro1993ezw}.
Whereas rendering and compression operate on existing scene or image content, generation can begin with a specified distribution of detail before the final pixels are available.
Notably, such knowledge can come from the structure, description, or plan of the intended scene, which generative models increasingly support through richer prompting and explicit layout specification~\citep{gptimage2,reve2,ideogram2026structuredprompts}.
These observations motivate using prior knowledge of where detail matters to determine how to allocate compute for generation.

Ideally, a generative model would use this prior knowledge both to adapt where fine detail is concentrated and to reduce computation. Spatial conditioning methods~\citep{zhang2023controlnet,t2i-adapter,ominicontrol,creatilayout,reco} guide generated content using spatial inputs, but retain full-resolution processing and often introduce additional computational cost. Efficiency methods exploit redundancy through either reusing intermediate computations~\citep{teacache,toca} or token reduction~\citep{bolya2023token,lu2025toma,xiao2026spectral,ralu,kim2026ddit,wu2025crpa,chao2026foveateddiffusion}, but do not utilize this adaptivity as a control mechanism.
Our goal is to use prior knowledge to control where fine detail is concentrated and allocate computation accordingly, enabling both adaptive and efficient generation.

In this paper, we introduce \textit{Level-of-Token (LoT) Diffusion}, in which an image or video is synthesized from a content-adaptive, multiresolution token layout (\textit{Level-of-Token Layout}). Each token represents a rectangular spatial patch with variable extent, specified by its height and width, allowing important regions to be finely tokenized while less important ones are tokenized coarsely.
Realizing this on pretrained diffusion transformers requires adapting models trained on uniform grids to heterogeneous token layouts. 
To this end, we develop a formulation that preserves full-resolution flow prediction at every denoising step while the transformer processes only a reduced token sequence. We achieve this through a patch-wise asymmetric flow parameterization~\citep{chen2026asymmetric}, together with adaptations to token embeddings and positional encoding that account for various patch extents. Fine-tuning pretrained models within our simple framework enables efficient generation that adapts to the specified layout while leveraging their generative priors.

We demonstrate LoT across diverse layout sources, including semantic masks, bounding boxes, texture variance, and depth-of-field cues, as well as agentic plans. Generation speedup depends on the token budget specified by the layout, reaching up to $2.04\times$ for images and $3.53\times$ for videos on average relative to the full-resolution pretrained models. Across the evaluated budgets, LoT offers favorable tradeoffs between generation quality and efficiency compared with existing token-reduction methods. Qualitative results further show how generated content adapts to the specified token layout.
Our project website is at \url{https://georgenakayama.github.io/lotdiffusion/}. 

To summarize, our main contributions are:
\begin{itemize}
    \itemsep0em
    \item We introduce \textit{Level-of-Token Diffusion}, a framework for efficient image and video generation with token layouts that adaptively vary patch size and shape across an image or video.
    \item We adapt pretrained DiTs through patch-wise asymmetric flow and embeddings for multiresolution tokens, enabling full-resolution flow prediction at every step from a reduced token sequence.
    \item We demonstrate favorable quality and efficiency tradeoffs across image and video generation, using layouts derived from scene cues and agentic plans.
\end{itemize}

\section{Related Work}

\paragraph{Methods for efficient generation.} A large body of work improves
the efficiency of diffusion models along several axes. Along the spatial
axis, token merging and pruning techniques identify redundant tokens and
merge or discard them based on predefined heuristics or learned criteria
\cite{bolya2023token, kim2024tokenfusion, lu2025toma, haurum2024agglomerative, wu2025importance, lee2025local, fang2025attend},
while sparse attention mechanisms exploit sparse patterns in the attention
maps and use block-wise computation to reduce the quadratic cost of
self-attention
\cite{li2025radial, xia2025trainingfree, zhan2025bidirectional, katharopoulos2020linear, wang2020linformer, choromanski2021performer, beltagy2020longformer}.
Along the timestep axis, feature caching methods reuse intermediate
features across denoising steps to avoid redundant computation
\cite{teacache, toca, freqca, ma2023deepcache}, whereas distillation and
sampler-based approaches reduce the number of denoising steps themselves,
either by training a student to approximate a pretrained teacher
\cite{dmd, dmd2, causvid, lcm, pcm, prog-dist} or by deriving faster
solvers and consistency-based samplers \cite{ddim, song2023consistency}.
Other progressive-upsampling methods grow the spatial
resolution over the course of generation to save computation in early
stages \cite{ralu, xiao2026spectral}. However, these methods treat
spatial regions equally, operating along denoising-trajectory or
attention-level axes independent of how details are
spatially distributed, and thus leave the prior
knowledge of \emph{where} content requires computation unexploited.

Content-adaptive tokenization
is an orthogonal and complementary axis that exploits exactly this prior,
accelerating generation by allocating compute based on the content. However,
existing approaches address only part of the problem: some are spatially
adaptive but with no or limited user control \cite{chao2026foveateddiffusion, dc-dit, dydit, zamfir2026hilbertpixel, dydit++},
others adapt the token budget without redistributing it across space
\cite{elastictok, flextok, infotok, cat2025, alit2025, elit}, and
still others achieve spatial adaptivity but require training from scratch and are incompatible with pretrained diffusion models
\cite{gpstoken, d2it, diff-hier-tok}. In contrast, our method is spatially
adaptive and compatible with various pretrained models, redistributing tokens
non-uniformly according to an intended layout while additionally exposing user control
over the allocation. Most importantly, our method greatly accelerates image and video generation while remaining compatible
with the other efficiency techniques above. 

\paragraph{Layout control for diffusion models.} A large body of work
augments diffusion models with explicit control over object placement and
scene composition. Prior conditioning-based methods attach trainable modules onto
a frozen backbone, injecting spatial signals such as depth,
segmentation masks, or bounding boxes, so that users can dictate where content appears
\cite{zhang2023controlnet, t2i-adapter}. More recently, another line of work makes layout a direct input to the model rather than a separate branch, encoding it into tokens that enter the denoising sequence
\cite{ominicontrol, omnigen, creatilayout} or expressing it as structured
text in the prompt itself \cite{reco}. Most strikingly, the strongest
deployed systems have made layout their primary interface: Ideogram 4.0 is
trained exclusively on structured JSON captions carrying a description and
a bounding box per element \cite{ideogram4}, and Reve 2 derives a
hierarchical layout from the input text prompt before rendering
from it \cite{reve2}. However, in all these methods the layout merely decides \textit{what} appears \textit{where}, never how much compute is allocated to it. A region the layout
identifies as empty sky is still generated on the same dense token grid as
detailed typography. 
In contrast, our method uses this prior knowledge of where detail matters to allocate computation spatially, enabling content-adaptive generation with fewer tokens. Semantic masks and depth-of-field cues identify regions that can be generated more coarsely, allowing LoT to reduce computation there while preserving finer tokenization where detail is needed.

\section{Preliminaries}
\label{sec:background}
We define level-of-token layouts and review flow-based image and video generation.
We denote the full-resolution latent grid by
$\bm{x}\in \mathbb{R}^{H_0\times W_0\times D'}$, where $D'$ is the
channel dimension. The concepts naturally extend to videos by
adding a temporal axis.

\subsection{Level-of-Token Layout}
\label{sec:bg:tokenization}

Existing image and video generative models represent visual content as
sequences of latent tokens. A pretrained DiT uses a fixed size
$(p_h,p_w)$ to rearrange $\bm{x}$ into a \emph{uniform token layout}:
\begin{equation}
\label{eq:uniform_tokenization}
\bm{x}_U=\operatorname{Patchify}(\bm{x})\in\mathbb{R}^{H\times W\times D},
\qquad H=H_0/p_h,\quad W=W_0/p_w,\quad D=p_hp_wD'
\end{equation}
where $\operatorname{Patchify}$ flattens each non-overlapping $p_h\times p_w$ latent patch into a $D$-dimensional token. 

Motivated by level-of-detail geometry, variable-rate shading, and adaptive image and video representations~\cite{clark1976lod,vaidyanathan2014coarse,shapiro1993ezw,infotok}, which allocate more budget to regions that are visually important or contain high-frequency textures, we generalize uniform token layouts to \emph{level-of-token (LoT) layouts}.
Formally, a LoT layout allows the patch's shape to vary spatially based on a partition
\(
  \mathcal{P}=\{R_1,\dots,R_L\}
\)
of the uniform token grid $[0,H)\times[0,W)$ into $L$ disjoint axis-aligned tokens:
\begin{equation}
\begin{gathered}
R_i = [u_i,\,u_i+e_{h,i}) \times [v_i,\,v_i+e_{w,i}), \\
0 \le u_i < u_i+e_{h,i} \le H, \qquad
0 \le v_i < v_i+e_{w,i} \le W,
\end{gathered}
\end{equation}
where $R_i$'s are pair-wise disjoint and cover the $H\times W$ token grid.
We denote the $i$-th token's extent as
$\bm e_i=(e_{h,i},e_{w,i})$ and define
$D_{\bm{e}_i}:=e_{h,i}e_{w,i}D$. Furthermore, let $\bm{x}^{(i)}\in\mathbb{R}^{D_{\bm e_i}}$ denote the
concatenation of all uniform tokens within region $R_i$ and let $\bar{\bm{x}}_{R_i}\in\mathbb{R}^{D}$ denote the single token in the LoT layout representing the spatial region of this patch. We will discuss the mapping between $\bm{x}^{(i)}$ and $\bar{\bm{x}}_{R_i}$ in Sec.~\ref{sec:method:asymflow}.

\begin{wrapfigure}{r}{0.4\textwidth} %
  \centering
  \vspace{-13pt}
  \vspace{-1\baselineskip} %
  \includegraphics[width=\linewidth]{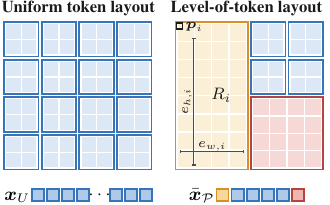}
  \vspace{-2.0\baselineskip} %
  \caption{\textbf{Token layout notation.}}
  \label{fig:tokenization}
    \vspace{-1.0\baselineskip} %
\end{wrapfigure}
The partition induces the level-of-token sequence $\bar{\bm{x}}_{\mathcal{P}}=(\bar{\bm{x}}_{R_1},\dots,\bar{\bm{x}}_{R_L})\in\mathbb{R}^{L\times D}$ (see Fig.~\ref{fig:tokenization}). We refer to its $i$-th token's top-left coordinate as $\bm{p}_i = (u_i, v_i)$. The same notation extends to videos by adding a temporal
axis where a latent grid of size $T_0\times H_0\times W_0$ is patchified into a
$T\times H\times W$ token grid, with $T=T_0/p_t$ and
$D=p_tp_hp_wD'$. Throughout the paper we refer to token compression relative to
the pretrained DiT as $HW/L$ for images and $THW/L$ for videos.

\subsection{Flow-based Image and Video Generation}
\label{sec:bg:flow}
\paragraph{Flow matching.}
Flow matching defines a probability path between the data distribution and a standard Gaussian noise distribution. This path couples a data sample $\bm{x}_0\in \mathbb{R}^{H_0\times W_0 \times D'}$ with a noise sample $\bm{x}_1 := \bm{\varepsilon} \sim \mathcal{N}(0, \bm{I})$ using the time schedules $\alpha_t = 1 - t$ and $\sigma_t = t$ for $t\in[0,1]$. The state and velocity remain on this full-resolution latent grid. We define the probability path and its velocity for any $t\in [0, 1]$ as
\begin{equation}
\bm{x}_t = \alpha_t \bm{x}_0 + \sigma_t \bm{\varepsilon}, \quad \bm{u} = \bm{\varepsilon} - \bm{x}_0 = \frac{\bm{x}_t - \bm{x}_0}{\sigma_t}.
\end{equation}
The flow model $v_\theta(\bm{x}_t, t, \bm{c})$ is trained to predict the flow velocity $\bm{u}$ given $\bm{x}_t$, time step $t$, and optional text conditioning $\bm{c}$, by optimizing the following loss:
\begin{equation}
\label{eq:flow-loss}
\mathcal{L}_{\mathrm{flow}}(\theta)
\;=\;
\mathbb{E}_{t,\bm{x}_0,\bm{x}_1, \bm{c}}\!\left[
\big\lVert v_\theta(\bm{x}_t,t,\bm{c}) - \bm{u}\big\rVert_{2}^{2}
\right].
\end{equation}
After training, generation solves the ODE $\frac{d\bm{x}_t}{dt} = v_\theta$ with $\bm{x}_1 = \bm{\varepsilon} \sim \mathcal{N}(\bm{0}, \bm{I})$ from $t=1$ to $0$.

\paragraph{Asymmetric flow matching.}
Recently, asymmetric flow matching (AsymFlow)~\cite{chen2026asymmetric} proposes to fine-tune pretrained image latent flow matching models by adopting an asymmetric parameterization where the model learns to predict the flow velocity from low-rank noise to full-rank data 
$\bm{u}_{\text{A}} = \bm{P}\bm{\varepsilon} - \bm{x}$ where $\bm{P} = \bm{A}\bm{A}^T\in \mathbb{R}^{D\times D}, \bm{A} \in \mathbb{R}^{D \times d}, D \geq d$ is a fitted orthonormal projector. The full-rank velocity can be recovered by 
\begin{equation}
  \bm{u} = \bm{P}\bm{u}_{\text{A}} + (\bm{I} - \bm{P})\frac{\bm{x}_t + u_{\text{A}}}{\sigma_t}.
\end{equation}
AsymFlow generates images directly in pixel space at a computational cost similar to that of the original latent flow models. 
Inspired by this, Level-of-Token generates a full-resolution latent grid from a level-of-token layout, enabling efficient, content-aware generation.

\paragraph{Diffusion Transformers (DiT) and RoPE.}
Under a uniform token layout in Eq.~\ref{eq:uniform_tokenization}, Diffusion Transformers (DiT)~\cite{peebles2023dit} embed the $HW$ tokens of $\bm{x}_t$ into the hidden dimension $d_{\mathrm{model}}$ and process them with transformer blocks. The output head predicts full-resolution velocities, which are rearranged into a full-resolution velocity field of size $H_0\times W_0\times D'$. Because self-attention is permutation equivariant, models such as FLUX.2 and Wan encode each token's spatial position $\bm{p}_i=(i_1,\dots,i_d)$ on the uniform token grid using multi-axis RoPE~\cite{su2024roformer}. For query and key vectors $\bm{q}_i,\bm{k}_i\in\mathbb{R}^{d_{\mathrm{h}}}$, where $d_{\mathrm{h}}$ is the attention-head dimension, RoPE splits the channels across the $d$ spatial axes and applies 
\begin{equation}
\label{eq:rope-axial}
\widetilde{\bm{q}}_i
=\mathbf{R}_{\mathrm{axial}}(\bm{p}_i)\bm{q}_i,
\qquad
\widetilde{\bm{k}}_i
=\mathbf{R}_{\mathrm{axial}}(\bm{p}_i)\bm{k}_i,
\qquad
\mathbf{R}_{\mathrm{axial}}(\bm{p}_i)
=\bigoplus_{a=1}^{d}\mathbf{R}^{(a)}(i_a).
\end{equation}

where $\mathbf{R}^{(a)}(i_a)$ denotes the standard RoPE rotation~\cite{su2024roformer} applied to the channel block assigned to spatial axis $a$. Usually, RoPE assigns a spatial position to each token and does not
encode the size of its spatial support. A DiT operating on a level-of-token layout must
therefore place all tokens in a consistent coordinate system and further encode
their extents separately.

\begin{figure}[t!]
    \centering
    \includegraphics[width=\textwidth,trim=0 3bp 0 0,clip]{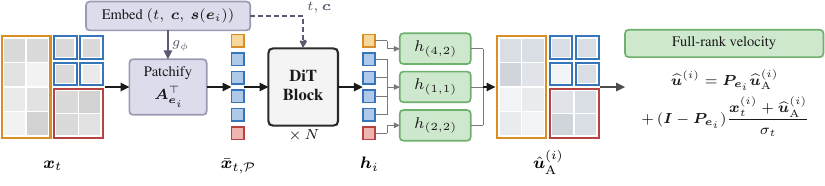}
    \caption{\textbf{Level-of-Token Diffusion pipeline.} LoT Diffusion maps a full-resolution token grid $\bm{x}_t$ to a sequence of reduced tokens specified by the LoT layout. The DiT blocks process these tokens conditioned on token shapes $\bm{s}$, timestep $t$, and text $\bm{c}$. Extent-dependent heads $h_{\bm{e}_i}$ restore the asymmetric velocity $\hat{\bm{u}}$, which is converted into the full-rank velocity for flow-matching training.}
    \label{fig:pipeline}
\end{figure}

\section{Method}
\label{sec:method}

Given a LoT layout defined by $\mathcal{P}=\{R_1,\ldots,R_L\}$, our goal is to
reduce the number of tokens processed by the DiT while remaining compatible with pretrained models. 
Specifically, we use patch-wise asymmetric flow matching 
(Sec.~\ref{sec:method:asymflow}) and introduce Level-of-Token DiT (Sec.~\ref{sec:method:dit}) that minimally modifies the DiT architecture to handle LoT layouts. We present our method for image generation, with a natural extension to video through per-frame LoT layouts. Fig.~\ref{fig:pipeline} shows our overall pipeline. 

\subsection{Flow Matching with Level-of-Tokens}
\label{sec:method:asymflow}

A LoT layout reduces the DiT sequence length from $HW$ to $L$,
but adapting a pretrained latent diffusion model to this reduced
sequence presents two challenges. First, the pretrained VAE decoder
requires a full-resolution latent grid of size
$H_0\times W_0\times D'$. Second, the pretrained DiT predicts flow
on the same sequence it receives, so directly applying it to
the reduced sequence would only produce $L$ token-level flow predictions, rather than the full-resolution flow
needed for denoising. 

\begin{figure}[t!]
    \centering
    \includegraphics[width=\linewidth]{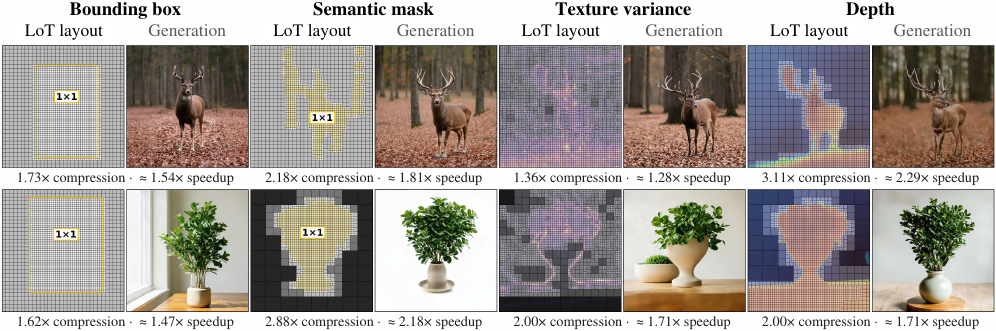}
    \caption{\textbf{Image generation with different layouts.} We generate images using arbitrary LoT layouts derived from
signals that indicate spatial importance, including bounding boxes, semantic masks, texture variance, or depth maps, allocating finer tokens to regions that require detail.}
    \label{fig:grid_layout}
\end{figure}

\paragraph{Patch-wise asymmetric flow.}
Inspired by AsymFlow~\cite{chen2026asymmetric}, we use compressed
LoT tokens to predict a dense asymmetric velocity field. For each extent $\bm{e}=(e_h,e_w)$, we fit a semi-orthonormal
basis $\bm{A}_{\bm{e}}\in\mathbb{R}^{D_{\bm{e}}\times D}$ using orthogonal Procrustes~\cite{Schönemann_1966} between aligned dense patches and extent-matched multi-scale VAE tokens
(see Sec.~\ref{sec:supp:fitting-scaling} for fitting details).
Given the noisy latent grid $\bm{x}_t$ and a LoT layout
$\mathcal{P}$, we project each region's patch onto its
corresponding basis and assemble the resulting token sequence:

\begin{equation}
\label{eq:method-projected-token}
    \bar{\bm{x}}_{t,R_i}
    =\bm{A}_{\bm{e}_i}^\top\bm{x}_t^{(i)}
    \in\mathbb{R}^{D},
    \qquad
    \bar{\bm{x}}_{t, \mathcal{P}}
    =\big(\bar{\bm{x}}_{t,R_1},\ldots,
           \bar{\bm{x}}_{t,R_L}\big)
    \in\mathbb{R}^{L\times D}.
\end{equation}
Thus, the DiT input dimension is the same as in the pretrained model, while its sequence length is determined by $\mathcal{P}$. For each region $R_i$ with extent $\bm{e}_i$, we further define $\bm{P}_{\bm{e}_i}=\bm{A}_{\bm{e}_i}\bm{A}_{\bm{e}_i}^\top$ to be the extent's projector. Then the dense asymmetric velocity target for $R_i$ becomes
\begin{equation}
\label{eq:method-asym-target}
    \bm{u}_{\mathrm{A}}^{(i)}
    =\bm{P}_{\bm{e}_i}\bm{\varepsilon}^{(i)}-\bm{x}_0^{(i)}.
\end{equation}
Given a prediction $\widehat{\bm{u}}_{\mathrm{A}}^{(i)}$, we recover a
full-rank velocity for the patch as
\begin{equation}
\label{eq:method-asym-recovery}
    \widehat{\bm{u}}^{(i)}
    =\bm{P}_{\bm{e}_i}\widehat{\bm{u}}_{\mathrm{A}}^{(i)}
    +(\bm{I}-\bm{P}_{\bm{e}_i})
      \frac{\bm{x}_t^{(i)}+
            \widehat{\bm{u}}_{\mathrm{A}}^{(i)}}{\sigma_t}.
\end{equation}
In practice, we clamp the denominator to
$\max(\sigma_t,10^{-6})$ for numerical stability.
For $t>0$, Eq.~\ref{eq:method-asym-recovery}
recovers the dense flow $\bm{\varepsilon}-\bm{x}_0$ when the asymmetric
target is predicted exactly.
We convert the recovered velocity into a clean-data estimate and train
using the loss in Eq.~\ref{eq:method-asym-loss}.

\paragraph{Extent-dependent output heads.}
A token of extent $\bm{e}$ must predict an asymmetric velocity for all
$e_he_w$ tokens in its patch. We therefore use one linear output head per supported extent $\bm{e}$,
\begin{equation}
\label{eq:method-resolution-head}
    h_{\bm{e}}:
    \mathbb{R}^{d_{\mathrm{model}}}
    \rightarrow\mathbb{R}^{D_{\bm{e}}},
    \qquad
    \widehat{\bm{u}}_{\mathrm{A}}^{(i)}
    =h_{\bm{e}_i}(\bm{h}_i) = \bm{W}_{\bm{e}}\bm{h}_i, \qquad \bm{W}_{\bm{e}}\in \mathbb{R}^{D_{\bm{e}}\times d_{\mathrm{model}}},
\end{equation}
where $d_{\mathrm{model}}$ is the DiT's hidden dimension. We apply Eq.~\ref{eq:method-asym-recovery} patch-wise to obtain a full-resolution velocity field, i.e., $\bm{u}\in\mathbb{R}^{H_0\times W_0\times D'}$. Furthermore, we initialize each output head as $\bm{W}_{\bm{e}}=\bm{A}_{\bm{e}}\bm{W}_{\mathrm{out}}$, where $\bm{W}_{\mathrm{out}}\in \mathbb{R}^{D\times d_{\mathrm{model}}}$ is the pretrained output head.

\paragraph{Extent-dependent data scaling.}
When fitting $\bm A_{\bm e}$, we also estimate a scalar $s_{\bm e}$ by matching projected and reference token
root-mean-square norms. AsymFlow~\cite{chen2026asymmetric} compensates for
this scale mismatch through input/output rescaling and timestep calibration.
However, applying that calibration per extent would introduce heterogeneous per-token
timesteps, departing from the pretrained DiT's shared-timestep conditioning.
Therefore, we instead calibrate at the data level, namely we divide each clean dense patch by $s_{\bm e}$ before adding noise and undo the scaling before VAE decoding.
Fitting and scaling details are provided in Sec.~\ref{sec:supp:fitting-scaling}
of the supplementary material.

\begin{figure}[t!]
    \centering
    \includegraphics[width=\linewidth]{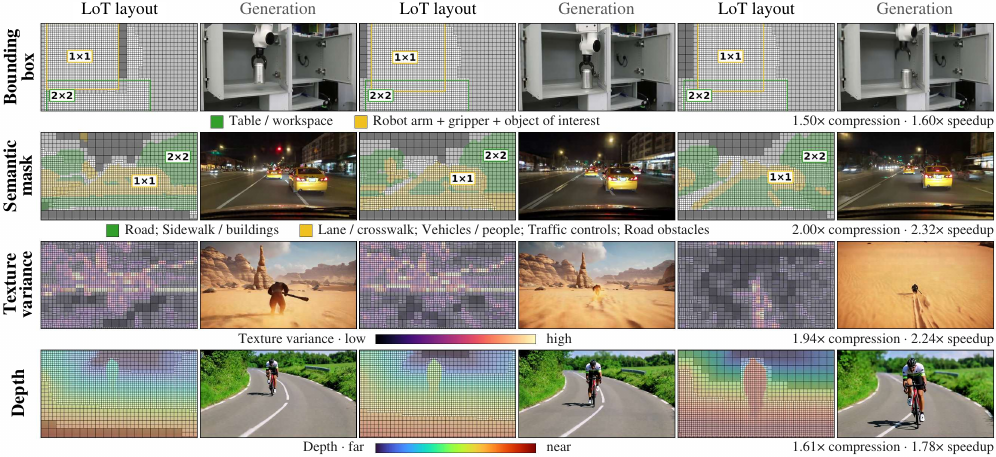}
    \caption{\textbf{Video generation with different layouts.} Our model generates video under arbitrary LoT layouts derived from bounding box, semantic mask, texture variance, or depth input, allocating tokens where
an application needs detail and reducing them elsewhere, such as sky and road
surface in autonomous driving and flat untextured surfaces in real-time
gaming.}
    \label{fig:grid_layout_video}
\end{figure}

\subsection{Level-of-Token DiT}
\label{sec:method:dit}

The projected tokens in Eq.~\ref{eq:method-projected-token} allows us to reuse the DiT’s 
pretrained weights. We further introduce extent-dependent input heads and initialize each by composing the pretrained input projection with $\bm A_{\bm e}^{\top}$. 
To inform the DiT of the Level-of-Token layout, we introduce patch-shape conditioning and finest-grid-aligned RoPE to encode each LoT token's spatial extent and location, respectively.

\paragraph{Patch-shape conditioning.}
To represent a token with extent
$\bm{e}_i=(e_{h,i},e_{w,i})$, we encode its per-axis size, area, and aspect
ratio as
\begin{equation}
\label{eq:method-size-features}
    \bm{s}(\bm{e}_i)
    =\big(
       \log_2 e_{h,i},\,
       \log_2 e_{w,i},\,
       \log_2(e_{h,i}e_{w,i}),\,
       \log_2(e_{h,i}/e_{w,i})
      \big).
\end{equation}
A zero-initialized MLP maps these features to the DiT hidden dimension and is added
to the input token embedding:
\begin{equation}
\label{eq:method-size-embedding}
    \bm{h}_i^{0}
    =\bm{W}_{\mathrm{in},\bm{e}_i}\bm{x}_t^{(i)}
     +g_{\phi}\!\left(\bm{s}(\bm{e}_i)\right).
\end{equation}

\paragraph{Finest-grid-aligned RoPE.}
We retain the pretrained DiT's axial RoPE and index each token at its patch
center on the uniform token grid
\begin{equation}
\label{eq:method-center-coordinate}
    \bm{c}_i
    =\left(u_i+\frac{e_{h,i}-1}{2},\,
           v_i+\frac{e_{w,i}-1}{2}\right),
\end{equation}
where $(u_i, v_i)$ is the top-left coordinate of the $i$-th token. We use $\bm{c}_i$ in Eq.~\ref{eq:rope-axial} for both queries and keys, while
the shape embedding conveys patch extent. This parameter-free RoPE change supports
standard dense attention across all resolutions. For unit patches,
$\bm{c}_i=(u_i,v_i)$, recovering the pretrained positional encoding. Unlike multi-scale RoPE variants such as~\citet{wu2025crpa, ralu, chao2026foveateddiffusion} that require query-dependent position maps,
our simple approach is query-independent and applies unchanged to every supported extent.

\section{Results}
\noindent \textit{Experimental Setup.} For image generation, we fine-tune FLUX.2-klein-base-4B~\cite{flux-2-2025}
on Aesthetic-Train-V2~\cite{zhang2025diffusion4k} at $1024\times1024$,
following~\cite{chao2026foveateddiffusion}, with a random 95\%/5\% split for
training and held-out evaluation. We also fine-tune the 9B variant using
the separate training setup described in Sec.~\ref{sec:supp:training}.
Image results use 4B unless otherwise noted. For video generation, we fine-tune
Wan2.1-T2V-14B~\cite{wan2025wan} on a selected 100k-example subset of
Vchitect-T2V-Dataverse~\cite{fan2025vchitect}, using 81-frame clips at 720P. We reserve an additional 200 text-video pairs for held-out evaluation.

LoT-Flux supports extents $(e_h,e_w)$ with $e_h,e_w\in\{1,2,4,8\}$, while
LoT-Wan uses $(1,e_h,e_w)$ with $e_h,e_w\in\{1,2,4\}$. Models used for
layout-controlled generation are trained on mixed-source layouts derived
from semantic masks, object bounding boxes, reference-image texture
variance, and depth maps. We adapt the pretrained
transformer projections with LoRA~\cite{hu2022lora} and fully fine-tune the
extent-dependent input and output heads and patch-size embeddings.
See the supplementary material for detailed training recipes:
Sec.~\ref{sec:supp:training} for images and
Sec.~\ref{sec:supp:video-training} for videos.

\Needspace{16\baselineskip}
\subsection{Layout-Adaptive Image and Video Generation}
\begin{wrapfigure}{r}{0.47\textwidth}
    \centering
    \vspace{-13pt}
    \includegraphics[width=\linewidth]{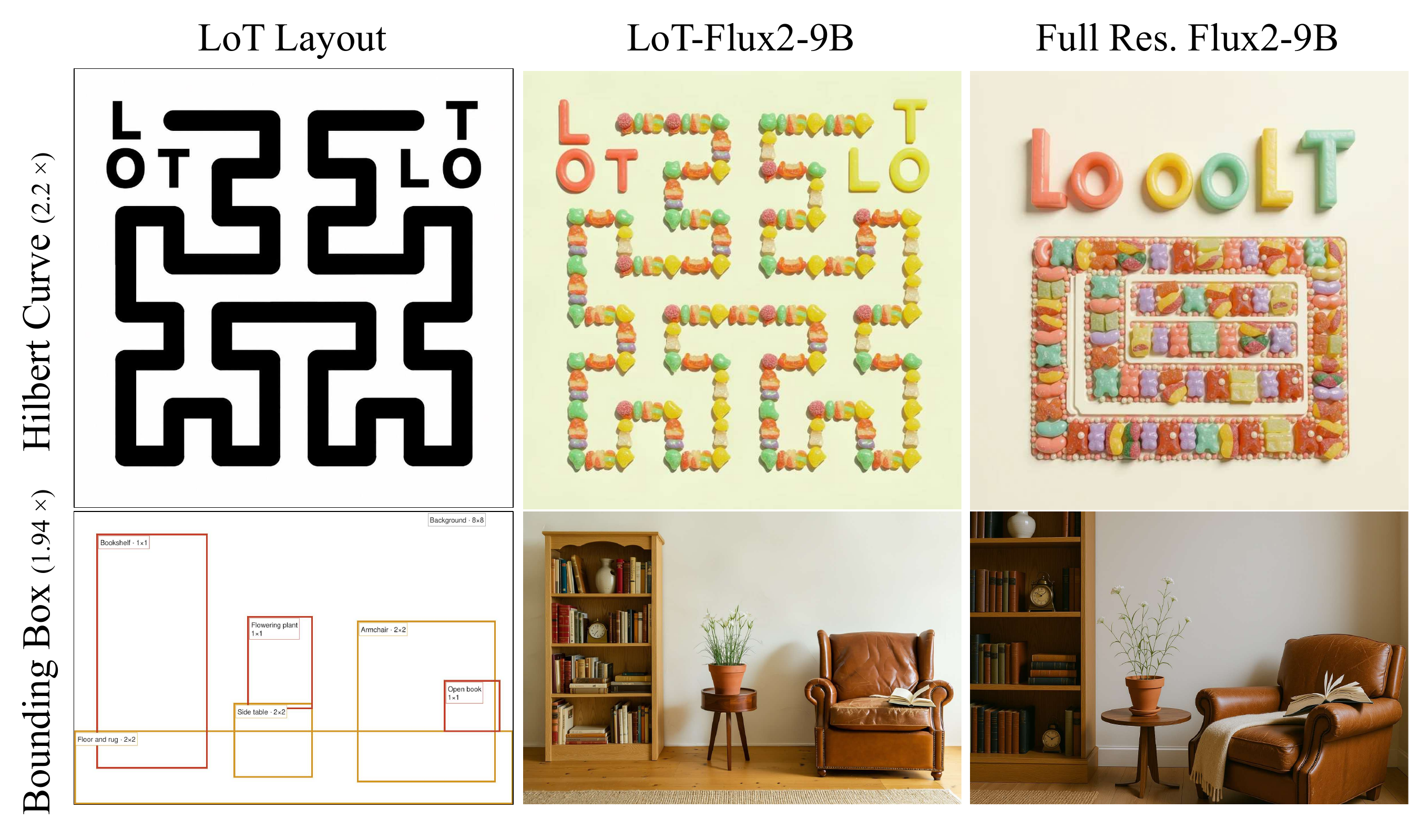}
    \vspace{-25pt}
    \caption{\textbf{Layout-adaptive image generation.}}
    \vspace{-13pt}
    \label{fig:user-layout-inset}
\end{wrapfigure}
Figures~\ref{fig:grid_layout} and~\ref{fig:grid_layout_video} show image and
video generation with token layouts derived from bounding boxes, semantic
masks, texture variance, and depth maps. Across these sources, LoT generates content that adapts to the token allocation, retaining intricate details in finer-token regions while favoring smoother or defocused content in coarser-token regions. Fig.~\ref{fig:user-layout-inset} further illustrates
this behavior with user-specified layouts using LoT-Flux2-9B.
For additional layout adaptive results see \href{https://georgenakayama.github.io/lotdiffusion/supplementary/}{supplementary website} and Sec.~\ref{sec:supp:layout-controllability}.

\begin{table*}[t!]
    \centering
    
    \resizebox{\textwidth}{!}{%
    \begin{tabular}{lccccccccc}
    \toprule
        Method & Compression & Speed & IR $\uparrow$ & HPSv2.1 $\uparrow$ & HPSv3 $\uparrow$ & FID $\downarrow$ & pFID $\downarrow$ & TOPIQ $\uparrow$ & MUSIQ $\uparrow$ \\
        \midrule
        Full Res. FLUX2 4B & 1.000$\times$ & 1.001$\times$ & 0.9518 & 0.2808 & 10.4299 & 16.748 & 20.884 & 0.5852 & 69.783 \\
        \midrule
        ToMe-SD                 & \multirow{4}{*}{1.893$\times$} & \underline{1.538$\times$} & 0.1757 & 0.2182 & 5.3569 & 25.757 & 30.502 & 0.4888 & 58.991 \\
        DDiT                 & & 1.340$\times$ & \textbf{0.8850} & 0.2613 & 9.0052 & 19.671 & 23.739 & 0.5593 & 66.233 \\
        Foveated Diffusion   & & 1.040$\times$ & 0.8280 & \underline{0.2615} & \underline{9.0092} & \underline{14.944} & \underline{18.595} & \underline{0.5822} & \underline{67.494} \\
        Ours (LoT-Flux2 4B)                & & \textbf{1.560$\times$} & \underline{0.8833} & \textbf{0.2728} & \textbf{10.1211} & \textbf{13.361} & \textbf{16.393} & \textbf{0.6080} & \textbf{69.691} \\
        \midrule
        ToMe-SD                 & \multirow{4}{*}{2.406$\times$} & \underline{1.793$\times$} & -0.2547 & 0.1946 & 2.9150 & 34.321 & 41.187 & 0.4488 & 53.558 \\
        DDiT                 & & 1.521$\times$ & \underline{0.7865} & 0.2512 & 8.0549 & 24.150 & 32.311 & \underline{0.5444} & 64.612 \\
        Foveated Diffusion   & & 1.100$\times$ & 0.7737 & \underline{0.2543} & \underline{8.2674} & \underline{15.702} & \underline{21.456} & 0.5433 & \underline{65.109} \\
        Ours (LoT-Flux2 4B)                & & \textbf{1.818$\times$} & \textbf{0.8513} & \textbf{0.2672} & \textbf{9.7533} & \textbf{13.308} & \textbf{17.046} & \textbf{0.5978} & \textbf{69.420} \\
        \midrule
        ToMe-SD                 & \multirow{4}{*}{2.935$\times$} & \underline{2.003$\times$} & -0.6029 & 0.1781 & 0.9253 & 43.528 & 50.116 & 0.4203 & 49.119 \\
        DDiT                 & & 1.321$\times$ & 0.7156 & 0.2447 & 7.4075 & 28.156 & 38.974 & \underline{0.5341} & \underline{63.435} \\
        Foveated Diffusion   & & 1.180$\times$ & \underline{0.7582} & \underline{0.2507} & \underline{7.8433} & \underline{16.124} & \underline{23.645} & 0.5097 & 63.105 \\
        Ours (LoT-Flux2 4B)                & & \textbf{2.040$\times$} & \textbf{0.8326} & \textbf{0.2639} & \textbf{9.5522} & \textbf{13.345} & \textbf{17.822} & \textbf{0.5873} & \textbf{68.760} \\
        \bottomrule
    \end{tabular}%
    }\vspace{-5pt}
    \caption{\textbf{Image generation baseline comparisons.} Our method achieves superior performance on most generation metrics across different token budgets, while remaining faster than the baselines.}
    \label{tab:image_gen_metrics}
\end{table*}

\begin{table*}[t]
    \centering
    \resizebox{\textwidth}{!}{%
    \begin{tabular}{lcccccccc}
        \toprule
        Method & Compression  & Speed $\uparrow$ & Aesthetic $\uparrow$ & Imaging $\uparrow$ & Dynamic $\uparrow$ & Background $\uparrow$ & Subject $\uparrow$ & Motion $\uparrow$ \\
        \midrule
        Full Res. Wan2.1 14B & 1$\times$ & 1.000$\times$ & 0.5518 & 0.6465 & 0.8900 & 0.9332 & 0.9254 & 0.9899 \\
        \midrule
        ToMe-SD & \multirow{3}{*}{2$\times$} & \underline{2.223$\times$} & 0.3497 & 0.4925 & 0.9000 & 0.9017 & 0.8235 & 0.9594 \\
        Foveated Diffusion & & 2.153$\times$ & \underline{0.5258} & \underline{0.5598} & \textbf{0.9650} & \textbf{0.9252} & \textbf{0.8800} & \textbf{0.9844} \\
        Ours (LoT-Wan2.1 14B) & & \textbf{2.262$\times$} & \textbf{0.5324} & \textbf{0.6120} & \textbf{0.9650} & \underline{0.9242} & \underline{0.8775} & \underline{0.9836} \\
        \midrule
        ToMe-SD & \multirow{3}{*}{2.5$\times$} & \underline{2.802$\times$} & 0.2930 & 0.4477 & 0.9300 & 0.9122 & 0.8071 & 0.9555 \\
        Foveated Diffusion & & 2.746$\times$ & \underline{0.5153} & \underline{0.5263} & \underline{0.9550} & \textbf{0.9236} & \textbf{0.8776} & \textbf{0.9839} \\
        Ours (LoT-Wan2.1 14B) & & \textbf{2.909$\times$} & \textbf{0.5204} & \textbf{0.5851} & \textbf{0.9650} & \underline{0.9232} & \underline{0.8729} & \underline{0.9832} \\
        \midrule
        ToMe-SD & \multirow{3}{*}{3$\times$} & \underline{3.305$\times$} & 0.2704 & 0.4208 & \textbf{0.9800} & 0.9189 & 0.8031 & 0.9526 \\
        Foveated Diffusion & & 3.276$\times$ & \underline{0.5080} & \underline{0.4952} & \underline{0.9550} & \textbf{0.9234} & \textbf{0.8761} & \textbf{0.9839} \\
        Ours (LoT-Wan2.1 14B) & & \textbf{3.530$\times$} & \textbf{0.5104} & \textbf{0.5650} & 0.9500 & \underline{0.9232} & \underline{0.8687} & \underline{0.9829} \\
        \bottomrule
    \end{tabular}%
    }
    \caption{\textbf{Video generation quantitative comparisons.} Our model obtains the best Aesthetic and Imaging scores compared with baselines and achieves the most speedup under the same budgets.}
    \label{tab:video_gen_metrics}
\end{table*}

\subsection{Runtime and Quality Analysis of Image and Video Generation}
We compare LoT with ToMe-SD~\cite{bolya2023token},
DDiT~\cite{kim2026ddit}, and Foveated Diffusion~\cite{chao2026foveateddiffusion}
for images, and with ToMe-SD and Foveated Diffusion for videos, using
LoT-Flux2-4B and LoT-Wan2.1-14B fine-tuned exclusively on texture-variance-derived layouts
and evaluating each method at matched token budgets. Speedups are measured
relative to the corresponding full-resolution model and cover generation
given a supplied layout, including diffusion sampling and VAE decoding and excluding layout
construction and model loading.
Comparison setups, metrics, and baseline details are in
Sections~\ref{sec:supp:image-comparison-details} and~\ref{sec:supp:video-comparison-details}
of the supplementary.

\begin{figure}[t!]
    \centering
    
    \includegraphics[width=\linewidth]{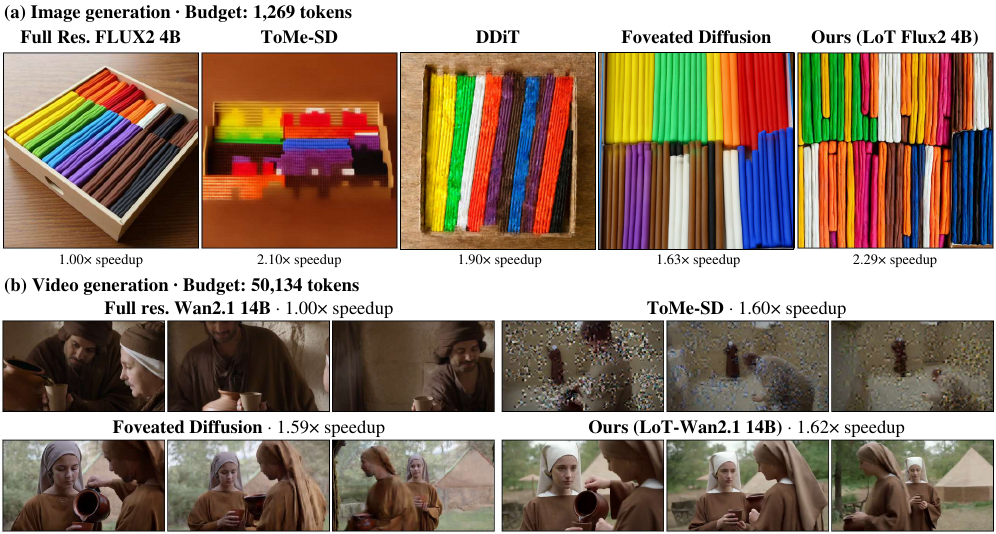}
    \caption{\textbf{Qualitative baseline comparisons} for (a) image and (b) video generation with the same token budget. ToMe-SD and DDiT lead to noisy low-res. regions. Foveated Diffusion results in distortions near mixed-res. token boundaries. We preserve visual detail while accelerating generation.}
    
    \label{fig:baseline_qualitative}
\end{figure}

For images, we assess predicted human preference using ImageReward~\cite{xu2023imagereward},
HPSv2.1~\cite{wu2023human}, and HPSv3~\cite{ma2025hpsv3widespectrumhumanpreference};
content fidelity using FID and pFID; and image quality using
TOPIQ~\cite{chen2024topiq} and MUSIQ~\cite{ke2021musiq}.
In Tab.~\ref{tab:image_gen_metrics}, LoT-Flux2-4B achieves
$1.32$--$2.04\times$ speedup and leads the baselines on most of the metrics across all token compression rates. For videos, Tab.~\ref{tab:video_gen_metrics} reports six
VBench scores on 200 held-out prompts.
LoT-Wan2.1-14B achieves $1.58$--$3.53\times$ speedup and the highest
aesthetic and imaging quality among the compressed methods at every
budget, while remaining competitive on the other evaluated metrics.
Qualitative image and video comparisons appear in
Figures~\ref{fig:baseline_qualitative}(a) and~\ref{fig:baseline_qualitative}(b), respectively. Additional image comparisons appear in Figures~\ref{fig:baseline_image_1} and~\ref{fig:baseline_image_2}.

\Needspace{15\baselineskip}
\subsection{Ablation Study}
\begin{wraptable}{r}{0.48\textwidth}
    \vspace{-\intextsep}
    \vspace{-2pt}
    \centering
    \small
    \setlength{\tabcolsep}{3pt}
    \resizebox{\linewidth}{!}{%
    \begin{tabular}{lccc}
        \toprule
        Variant & HPSv3 $\uparrow$ & FID $\downarrow$ & MUSIQ $\uparrow$ \\
        \midrule
        Direct HR prediction & 7.0595 & 26.487 & 63.346 \\
        Binary grid & 8.1955 & 20.603 & 66.443 \\
        No size embedding & \underline{10.0284} & \underline{14.584} & \underline{69.433} \\
        Ours & \textbf{10.3477} & \textbf{13.800} & \textbf{70.410}\\
        \bottomrule
    \end{tabular}%
    }
    \vspace{-1em}
    \vspace{-2pt}
    \caption{\textbf{Ablation study.}}
    \vspace{-1em}
    \label{tab:ablation_dit}
\end{wraptable}

Tab.~\ref{tab:ablation_dit} compares three variants of our framework design choices at a matched token budget: removing patch-size conditioning (\emph{No size embedding}), replacing patch-wise asymmetric flow recovery with a vanilla flow parameterization
(\emph{Direct HR prediction}), and restricting token extents to
$1\times1$ and $2\times2$ similar to previous works~\cite{chao2026foveateddiffusion} (\emph{Binary grid}). The full model performs
best on all metrics except ImageReward, with particularly large gains over
direct HR prediction and binary grids. Implementation details
and additional qualitative and quantitative results are provided in
Sec.~\ref{sec:supp:ablation} of the supplementary material.
\par

\section{Conclusion}
We introduced Level-of-Token Diffusion, a framework that uses level-of-detail information to guide efficient multiresolution image and video generation. By adapting pretrained diffusion models to variable token layouts, LoT connects control over detail with computational savings. Across diverse layout sources, our results demonstrate generation that adapts to the specified layout, with substantial speedups and favorable quality--efficiency tradeoffs over token-reduction baselines.
We note that scenes requiring fine structures or textures throughout offer limited opportunities for token reduction, and aggressive compression can degrade their fidelity. LoT is therefore most advantageous when detail requirements vary spatially, allowing computation to be reduced selectively. See Sec.~\ref{sec:supp:limitation} for a more detailed discussion.

Looking ahead, LoT suggests a closer connection between planning and efficient generation. Our demonstrations with agentic plans for images and videos provide an initial step, turning planned spatial importance into an explicit computational allocation. 
Level-of-Token layouts could serve as an editable intermediate representation, allowing multimodal systems and users to refine a visual plan and its computational budget.
Extending this interaction to video could support faster iterations between planning and synthesis in robotics and interactive simulation, where low generation latency is essential for responsive interaction.
More broadly, LoT offers a step toward generative systems that use their understanding of the intended content to decide where computation is needed.

\subsection*{Acknowledgment}
The work is supported by a Google unrestricted gift. Kiyohiro Nakayama is supported by a NSF GRFP fellowship; Brian Chao is supported by a NSF GRFP fellowship and a Stanford Graduate Fellowship. We also thank Mia Tang and Howard Xiao for their insightful conversation in the beginning of the project. This work used the following computation resources: Marlowe~\cite{marlowe2025}, Stanford University's GPU-based Computational Instrument, supported by Stanford HAI and Stanford Research Computing; NCSA DeltaAI at University of Illinois through allocation 261155 and 261171 from the Advanced Cyber infrastructure Coordination Ecosystem: Services and Support (ACCESS) program, which is supported by U.S. National Science Foundation grants \#2138259, \#2138286, \#2138307, \#2137603, and \#2138296; and a Modal Academic Grant. 

\bibliography{references}
\bibliographystyle{plainnat}

\clearpage
\appendix
\section{Appendix}

Unless otherwise specified, image-generation results in the main paper and
this appendix use LoT-FLUX.2-4B and the corresponding 4B baselines.
Figure~\ref{fig:user-layout-inset} uses LoT-FLUX.2 9B. Additional LoT-FLUX.2 9B
results are provided on the \href{https://georgenakayama.github.io/lotdiffusion/supplementary/}{supplementary website}.

\begingroup
\etocsettocstyle{\subsection*{Contents}}{}
\etocsetnexttocdepth{subsection}
\localtableofcontents
\endgroup

\subsection{Supplementary Evaluation Material}
\label{sec:supp:website}
The project website complements the paper and PDF appendix with
agent-assisted image and video generation, user-painted layouts, typography,
bounding-box-derived layouts, and layout-adaptive results from semantic masks,
bounding boxes, texture variance, and depth maps. The examples pair source
cues and token layouts with generated content; video playback illustrates how
allocation and generated detail evolve over time.
Website: \url{https://georgenakayama.github.io/lotdiffusion/supplementary/}.

\subsection{Extent-dependent Subspace Fitting and Data Scaling}\label{sec:supp:fitting-scaling}
For each extent $\bm e=(e_h,e_w)$, we collect aligned pairs
$(\bm x,\bar{\bm z})\sim\mathcal D_{\bm e}$. For both image and video
models, we fit the Procrustes bases and associated scales on a randomly
sampled subset of the respective training split. Here,
$\bm x\in\mathbb R^{D_{\bm e}}$ is a vectorized dense patch, and $\bar{\bm z}\in\mathbb R^D$ is the
corresponding token from an extent-matched multi-scale VAE encoding after
the pretrained model's fixed patchification. Thus, $D$ denotes the
patchified token dimension, and extents are measured on the uniform token
lattice, as in Sec.~\ref{sec:bg:tokenization}. Following
AsymFlow~\cite{chen2026asymmetric}, we fit a semi-orthonormal lift
$\bm A_{\bm e}\in\mathbb R^{D_{\bm e}\times D}$ with orthogonal Procrustes:
\begin{equation}
\label{eq:method-procrustes-objective}
    \bm A_{\bm e}^{\star}
    =\underset{\bm A^\top\bm A=\bm I_D}{\arg\min}\;
      \mathbb E_{(\bm x,\bar{\bm z})\sim\mathcal D_{\bm e}}
      \left[\left\|\bm x-\bm A\bar{\bm z}\right\|_2^2\right].
\end{equation}
The fitted lift defines the extent-specific projector
$\bm P_{\bm e}=\bm A_{\bm e}\bm A_{\bm e}^{\top}$.

Projection alone does not preserve the pretrained token magnitude. We
therefore fit a scalar data scale for each extent by matching projected-token
norms:
\begin{equation}
\label{eq:method-data-scale}
    s_{\bm e}
    ={\left(
      \frac{
        \mathbb E_{(\bm x,\bar{\bm z})\sim\mathcal D_{\bm e}}
        \left[\|\bm A_{\bm e}^{\top}\bm x\|_2^2\right]
      }{
        \mathbb E_{(\bm x,\bar{\bm z})\sim\mathcal D_{\bm e}}
        \left[\|\bar{\bm z}\|_2^2\right]
      }
    \right)}^{1/2}.
\end{equation}
Consequently, $\bm A_{\bm e}^{\top}\bm x/s_{\bm e}$ and
$\bar{\bm z}$ have the same root-mean-square norm. Given a partition
$\mathcal P$, we broadcast the corresponding scale over every dense site in
each patch and apply the patch-wise asymmetric flow
of Sec.~\ref{sec:method:asymflow} in the scaled data space:
\begin{equation}
\label{eq:method-scaled-data}
    s_{\mathcal P}(\bm q)=s_{\bm e_i}\quad\text{for }\bm q\in R_i,
    \qquad
    \bm y_0(\bm q)=\frac{\bm x_0(\bm q)}{s_{\mathcal P}(\bm q)},
    \qquad
    \bm y_t=\alpha_t\bm y_0+\sigma_t\bm\varepsilon.
\end{equation}
We use Eq.~\ref{eq:method-asym-target} and~\ref{eq:method-asym-recovery}
with $(\bm x_0,\bm x_t)$ replaced by $(\bm y_0,\bm y_t)$.
The model predicts asymmetric velocity, from which we recover the dense
velocity $\widehat{\bm u}$ and form the clean-data estimate
$\widehat{\bm y}_0=\bm y_t-\sigma_t\widehat{\bm u}$.
We train this velocity parameterization using a clean-data loss with
clamped noise-level weighting:
\begin{equation}
\label{eq:method-asym-loss}
    \mathcal L
    =\mathbb E\!\left[
      \frac{\left\|\widehat{\bm y}_0-\bm y_0\right\|_2^2}
           {\max(\sigma_t,0.05)^2}
      \right].
\end{equation}
At inference time, we integrate the flow in $\bm y$-space and map the result
back to the pretrained latent space as
$\widehat{\bm x}_0(\bm q)=s_{\mathcal P}(\bm q)
\widehat{\bm y}_0(\bm q)$. The multi-scale VAE tokens are used only to fit
$\bm A_{\bm e}$ and $s_{\bm e}$; both remain fixed during training and
inference.

\paragraph{Comparison with timestep calibration.}
For a single global scale $s$, AsymFlow~\cite{chen2026asymmetric} retains
the flow in the original data coordinates and compensates for the scale
mismatch through input/output rescaling and a calibrated internal timestep.
A patch-wise extension to our extent-dependent scales would use
\begin{equation}
\label{eq:supp:extent-timestep-calibration}
    k_{\bm e}(t)=\frac{1}{s_{\bm e}(1-t)+t},
    \qquad
    \tau_{\bm e}(t)=t\,k_{\bm e}(t).
\end{equation}
Token $i$ would then use input factor $k_{\bm e_i}(t)$ and internal timestep
$\tau_{\bm e_i}(t)$. When scales differ, these internal timesteps generally
differ within a sample, departing from the pretrained DiT's shared-timestep
conditioning. We do not use this extent-dependent timestep calibration.
Instead, we scale the clean data before adding standard Gaussian noise
as in Eq.~\ref{eq:method-scaled-data}. All tokens then use the same flow timestep
$t$, and the orthonormal projections preserve standard Gaussian projected
noise, avoiding this additional conditioning mismatch.

\subsection{Deriving Token Layouts from Different Sources}\label{sec:supp:layout-sources}
We describe how we derive level-of-token layouts from semantic masks and
bounding boxes, texture variance (VRS), depth maps, and custom detail maps. The same spatial
allocation rules apply to images and video frames.

\paragraph{Semantic masks and bounding boxes.}
During training, we sample up to three regions defined by semantic
masks or bounding boxes. Masks can also come from S3OD~\cite{kupyn2026s3od},
using the same token-allocation procedure.
Each selected region $M_k$ is assigned a token level
$\ell_k\in\{0,1,2,3\}$, corresponding to a spatial token extent
$b_k=2^{3-\ell_k}$ in each axis. Thus, levels $0,1,2,3$ correspond to
$8\times8$, $4\times4$, $2\times2$, and $1\times1$ tokens, respectively.
We also assign the background a level $\ell_{\mathrm{bg}}$ and sample
$\ell_k>\ell_{\mathrm{bg}}$, so selected regions receive finer tokens.
The resulting spatial assignment is
\begin{equation}
\label{eq:supp:region-token-levels}
    \ell(\bm q)=\max\!\left(
        \{\ell_{\mathrm{bg}}\}\cup
        \{\ell_k:\bm q\in M_k\}
    \right),
    \qquad b(\bm q)=2^{3-\ell(\bm q)}.
\end{equation}
This assigns a token level throughout each region and the background, with
the finer level taking precedence where selected regions overlap. We use
the resulting extent map to construct the level-of-token layout.

\paragraph{Texture variance (Variable-Rate Shading).}
We adapt the content-adaptive VRS rule of \citet{yang2019visually} to select
token extents using directional luminance variation. Candidate regions are
coarsened along directions whose estimated error falls below the tolerance
\begin{equation}
    \tau(R)=s\bigl(\mathbb E_R[I]+a\bigr),
\end{equation}
where $I$ is reference luminance. The sensitivity $s$ and ambient term $a$
control the overall detail of the token layout: larger values permit coarser
tokens, while smaller values preserve finer detail. For threshold-based VRS
training, we independently sample $s\in[0.01,0.1]$ and
$a\in[0.01,1.0]$ uniformly.
For quantitative evaluation, reference luminance comes from held-out
reference images for image comparisons and prompt-generated reference
videos for video comparisons, as detailed in
Sec.~\ref{sec:supp:image-comparison-details} and~\ref{sec:supp:video-comparison-details}.

\paragraph{Depth maps and depth of field.}
Following the circle-of-confusion model used in computational bokeh
rendering~\citep{wadhwa2018synthetic,yang2025any}, we convert a positive depth
map $d(\bm q)$ and focal depth $d_f$ into a blur-radius field
\begin{equation}
    r(\bm q)=K\left|d(\bm q)^{-1}-d_f^{-1}\right|.
\end{equation}
The focal depth is the median depth of a selected subject region. We calibrate
$K$ for the reference depth map and selected focal plane as
\begin{equation}
    K=\frac{r_{\mathrm{target}}}
    {\operatorname{percentile}_{95}\!\left(
        \left|d(\bm q)^{-1}-d_f^{-1}\right|
    \right)},
\end{equation}
where the percentile is computed over the reference pixels and $r_{\mathrm{target}}$
sets the desired blur strength. The first equation gives a small blur radius
$r(\bm q)$ when a pixel's depth is close to the focal depth $d_f$, and a
larger radius as its inverse depth differs from the focal plane. The second
equation scales these radii so their 95th percentile equals
$r_{\mathrm{target}}$. We then use this radius map to choose token sizes:
small radii require fine tokens, while larger radii permit coarser tokens.
A region is merged only when its smallest blur radius supports the larger
token size. Thus, even a small in-focus part prevents a region from being
coarsened, preserving detail around the subject and its boundaries.
Increasing $r_{\mathrm{target}}$ increases $K$ and the blur radii, allowing
coarser layouts.

\paragraph{Custom detail maps.}
Users can specify where to allocate computation by painting a spatial detail
map, without requiring a reference image or an automatically extracted mask.
An agent can also construct this map programmatically from a text description.
In our image-generation interface, each pixel is assigned the score of its
nearest brush color: black, orange, yellow, and white correspond to detail
scores $0$, $0.3$, $0.6$, and $1$, respectively. The canvas initially uses
yellow to provide medium detail throughout the image. We area-average the
scores onto the finest token lattice, optionally apply Gaussian smoothing,
and multiply by a user-controlled gain before clipping to $[0,1]$. This gives
a detail field $D(\bm q)$, where larger values request finer tokens.

We convert $D$ into a valid, lattice-aligned quadtree tiling. Starting from
$8\times8$ blocks, a block $R$ of extent $b\in\{8,4,2\}$ is split into four
children whenever
\begin{equation}
    \max_{\bm q\in R}D(\bm q)\geq t_b,
    \qquad t_8\leq t_4\leq t_2.
\end{equation}
Recursion stops when the test fails or a $1\times1$ token is reached. The
maximum is taken after projection onto the token lattice, so a high-detail
cell prevents its containing block from remaining coarse. The thresholds,
smoothing, and gain let the user adjust the resulting token count without
requiring an exact token budget or VRS-based refinement. The model receives
the resulting token layout and text prompt, not the painted RGB map as an
additional appearance condition. The image-generation demos on the project website
demonstrate both manual painting and agent-created custom detail maps.

\subsection{Image Generation Training Details}\label{sec:supp:training}

We fine-tune FLUX.2-klein-base-4B~\cite{flux-2-2025} with its standard latent
normalization, keeping the pretrained text encoder and VAE frozen.

\paragraph{Training data and source mixture.}
We train on Aesthetic-Train-V2~\cite{zhang2025diffusion4k}, using
aspect-preserving resizing and $1024\times1024$ center crops, with a random
95\%/5\% split for training and held-out evaluation. For each training image,
we derive its tokenization layout from segmentation masks, bounding boxes
from SAM3~\cite{sam3}, image-luminance variation (texture variance), and
depth maps~\cite{depthanything3}.

\paragraph{Training setup.}
We fully fine-tune the per-extent input/output heads, patch-size MLP, and
final adaptive normalization. We apply LoRA~\cite{hu2022lora} with rank
$r=256$ and dropout 0.05 to the image and context query/key/value and output
projections in the dual-stream blocks; image and context feed-forward
\texttt{linear\_in} and \texttt{linear\_out} layers; the fused
\texttt{to\_qkv\_mlp\_proj} inputs and single-stream attention
\texttt{to\_out} projections; and the timestep embedder. All other pretrained
parameters remain frozen.

We train for 50,000 steps using AdamW with a total batch size of 32,
learning rate $10^{-4}$, $(\beta_1,\beta_2)=(0.9,0.999)$, and $\epsilon=10^{-8}$.
Weight decay of $0.01$ is applied only to the LoRA adapters. We train the
asymmetric-velocity predictor with the clean-data loss
in Eq.~\ref{eq:method-asym-loss}, using the scaled space
of Eq.~\ref{eq:method-scaled-data}. The weight $1/\max(\sigma_t,0.05)^2$
is bounded near zero noise. We use
power-function EMA~\cite{Karras_2024_CVPR} with $\gamma=7$ for evaluation.

\paragraph{9B model.}
We also fine-tune FLUX.2-klein-base-9B~\cite{flux-2-2025}. We use three
million images from the LAION-3M dataset used by Asymmetric Flow
Models~\cite{chen2026asymmetric} for the semantic-mask,
bounding-box, saliency, and VRS branches, and bokeh-rendered images from
Aesthetic-Train-V2~\cite{zhang2025diffusion4k} for the depth-of-field branch.
The five branches are sampled with equal probability. We train on 32 NVIDIA
H100 GPUs across four nodes with a total batch size of 256, and use the
24,500-step EMA checkpoint for the qualitative results on the supplementary website.

\subsection{Image Generation Baseline Comparison Details}
\label{sec:supp:image-comparison-details}

\paragraph{Inference parameters.}
For all baselines, we generate with 30 sampling steps,
classifier-free guidance scale 4, and orthogonal guidance strength 1.
We evaluate on all 5,261 prompts in the held-out split.
For each prompt, we derive the VRS layout from its corresponding held-out
reference image. This is a layout-conditioned comparison: the layout
provides reference-derived spatial detail allocation beyond the text
prompt, but the reference RGB image is not an input to the generator.

\paragraph{Training Details.}
For a fair comparison, we extend DDiT~\cite{kim2026ddit} and Foveated
Diffusion~\cite{chao2026foveateddiffusion} to support token levels with spatial
extents up to $8\times8$.
For these comparisons, we train ours, DDiT, and Foveated Diffusion for
20,000 steps each exclusively on VRS-derived grids using the same settings:
rank-256 LoRA with dropout 0.05 on attention, feed-forward, and
timestep projection layers, AdamW with learning rate $10^{-4}$ and weight
decay $0.01$, and a global batch size of 16. We evaluate the 20,000-step
power-function EMA checkpoint with $\gamma=7$.

We also fine-tune a full-resolution FLUX.2-klein-base-4B control on the same
training split as our VRS-only quantitative model, using the same LoRA
target layers, rank 256, dropout 0.05, AdamW learning rate $10^{-4}$,
weight decay $0.01$, global batch size 16, and 20,000 training steps.

\paragraph{Metric details.}
We evaluate overall generation quality, content fidelity, image quality, and
efficiency. For overall generation quality, ImageReward
(IR)~\cite{xu2023imagereward}, HPSv2.1~\cite{wu2023human}, and
HPSv3~\cite{ma2025hpsv3widespectrumhumanpreference} are learned predictors of
human preference for prompt--image pairs, capturing prompt alignment and
visual appeal. Higher scores are better.

For content fidelity, Fr\'echet Inception Distance
(FID)~\cite{heusel2017gans} compares the feature distributions of generated
images and held-out reference images. Patch FID (pFID) applies the same
comparison to one native-resolution $299\times299$ center crop per image,
emphasizing local appearance.

For image quality, TOPIQ~\cite{chen2024topiq} combines semantic information
with local distortion cues, while MUSIQ~\cite{ke2021musiq} assesses
perceptual quality using multi-scale image representations. Both are used
without a reference image or text prompt; higher scores are better.

For efficiency, runtime measures the CUDA-synchronized generation call,
including prompt encoding, diffusion sampling, and VAE decoding, on an H100
per image; it excludes grid preprocessing, image saving, and model loading.
We retain the original pretrained full-resolution FLUX.2 4B timing reference
of $5.7783$ seconds per image for the reported speedups. The fine-tuned dense
control takes $5.7726$ seconds per image (steady-state mean), corresponding
to $1.001\times$ speed relative to that reference.

\subsection{Additional Image Generation Baseline Comparison Qualitative Results}
\begin{figure*}[!htbp]
    \centering
    \includegraphics[width=0.95\textwidth]{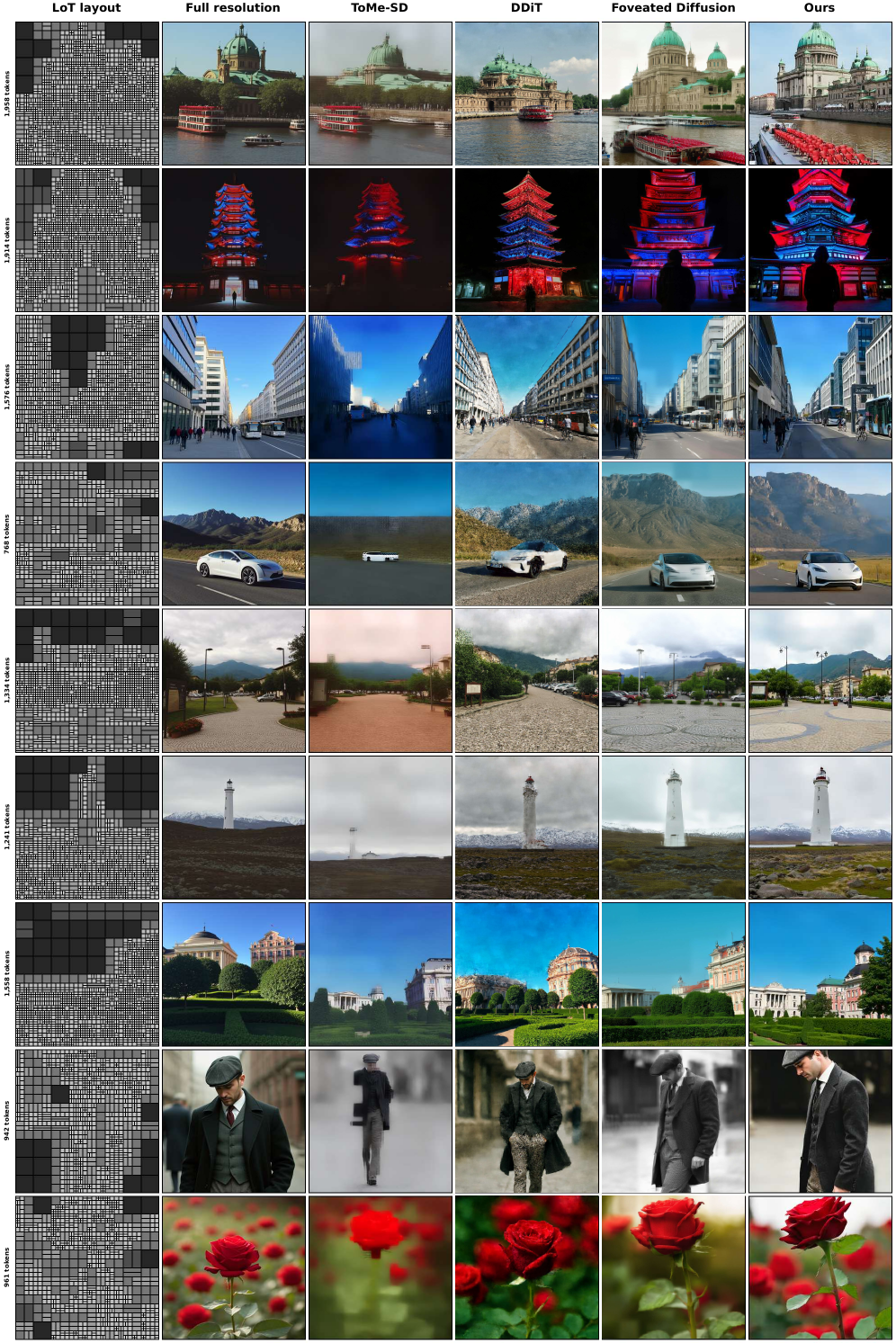}
    \caption{\textbf{Additional Qualitative Image Generation Baseline Comparison Results.}}\label{fig:baseline_image_1}
\end{figure*}

\begin{figure*}[!htbp]
    \centering
    \includegraphics[width=0.95\textwidth]{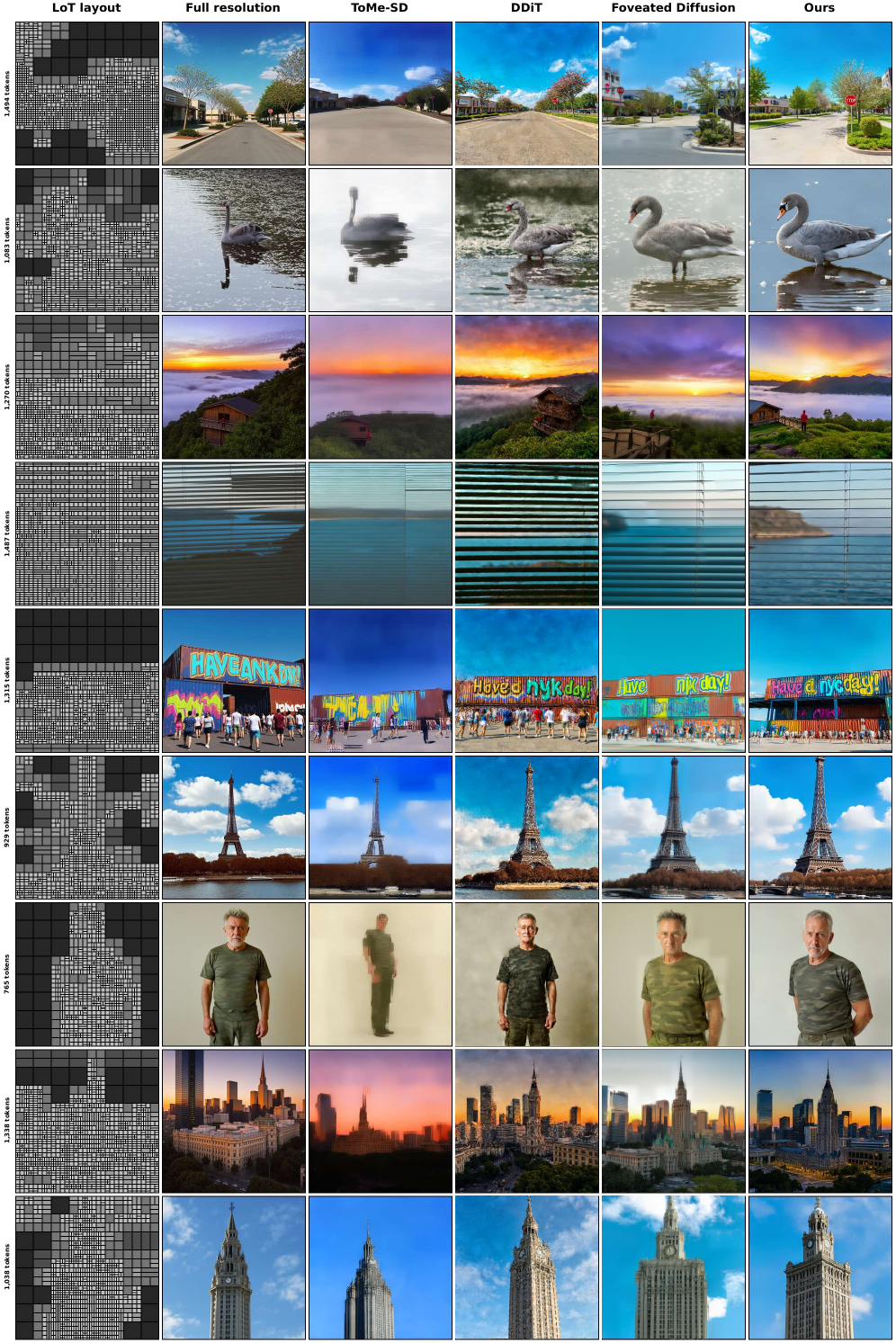}
    \caption{\textbf{Additional Qualitative Image Generation Baseline Comparison Results.}}\label{fig:baseline_image_2}
\end{figure*}

We show additional image generation baseline comparison qualitative results in Fig.~\ref{fig:baseline_image_1} and~\ref{fig:baseline_image_2}\@.

\FloatBarrier
\subsection{Additional Quantitative Comparison: Token Budget and Image Quality}
We provide an additional comparison between LoT-FLUX.2-4B and pretrained
FLUX.2 4B at matched per-image token budgets, using all 5,261 held-out test
prompts. We denote the pretrained baseline as \emph{FLUX.2 (token matched)}:
for each prompt, it uses a uniform square token grid whose token count is
as close as possible to that of our content-adaptive grid. It generates
at the corresponding lower resolution and is bicubically upsampled to
$1024\times1024$ for evaluation. LoT-FLUX.2-4B instead generates directly
at $1024\times1024$ using spatially varying token extents.

\begin{figure*}[!htbp]
    \centering
    \includegraphics[width=\textwidth]{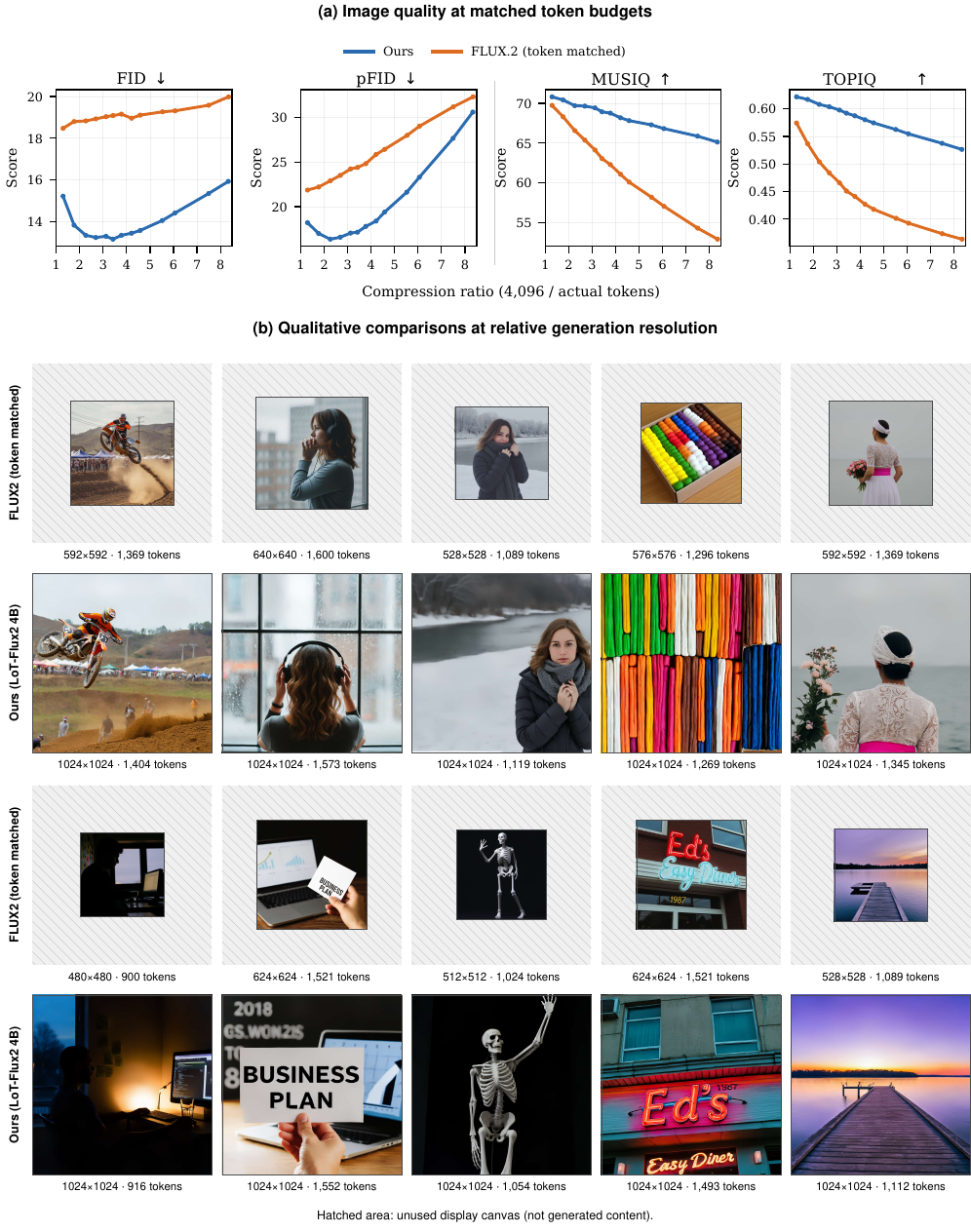}
    \caption{\textbf{Image Quality at Matched Token Budgets.} (a) Quantitative comparison. (b) Ten qualitative pairs, with token-matched outputs shown at their relative generation resolution. Hatched areas are display padding, not generated content.}\label{fig:tradeoff}
\end{figure*}

Fig.~\ref{fig:tradeoff}(a) compares FID and pFID against the held-out reference
images, together with the image-quality metrics MUSIQ and TOPIQ, averaged
over the test images at each threshold. Overall, LoT-FLUX.2-4B achieves lower
FID and pFID and higher image-quality scores than FLUX.2 (token matched).
This advantage is consistent with our adaptive allocation of finer tokens
to detailed regions while retaining full-resolution output. In contrast,
the pretrained baseline reduces resolution uniformly, and subsequent
upsampling cannot recover fine details lost during low-resolution generation.
Fig.~\ref{fig:tradeoff}(b) provides qualitative comparisons, with each
token-matched output displayed at its generation scale relative to our
$1024\times1024$ output.

\begin{figure*}[!htbp]
    \centering
    \includegraphics[width=\textwidth]{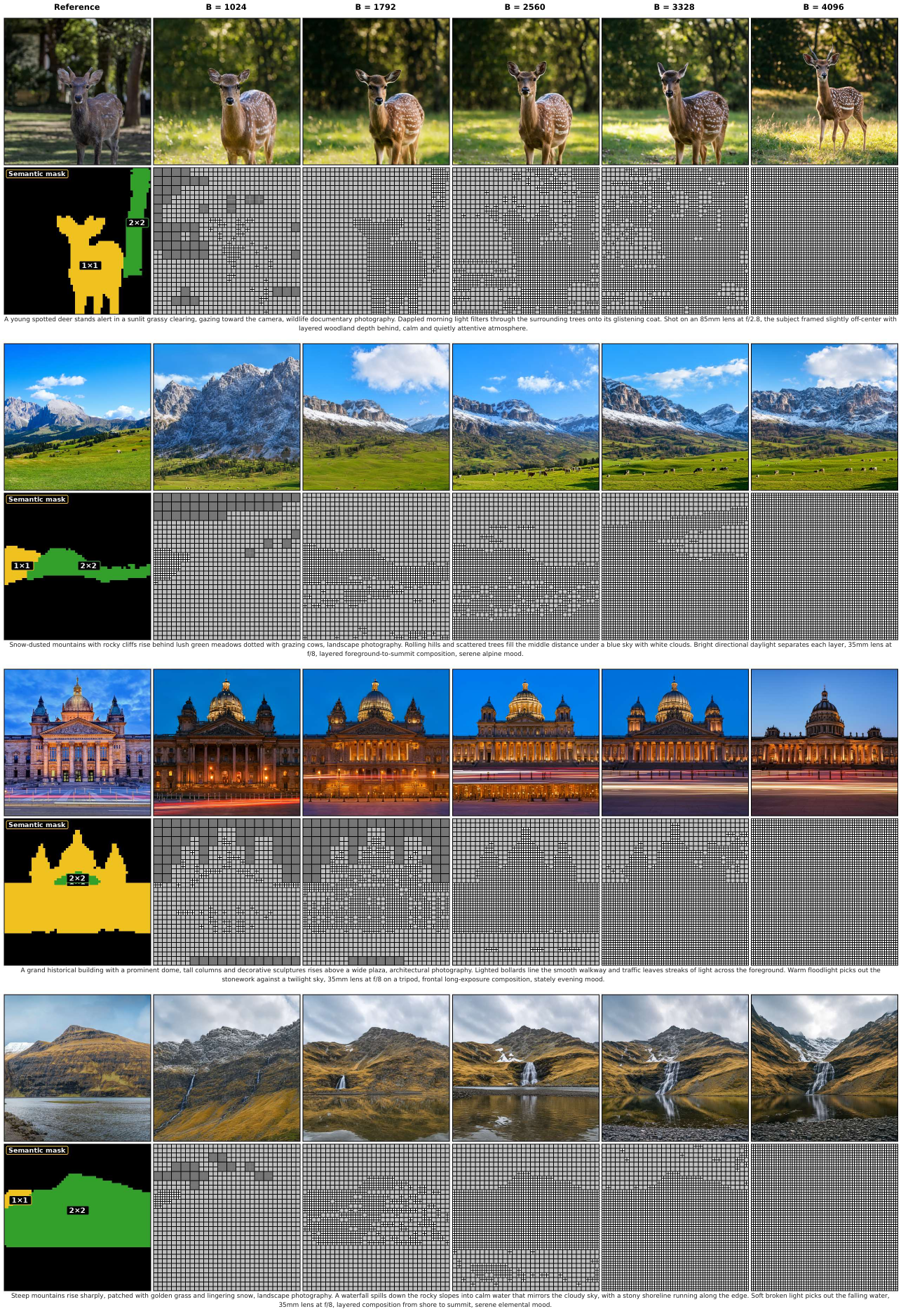}
    \caption{\textbf{Mask-guided Image Generation with Varying Token Budget.}}\label{fig:supp_budget_mask_1}
\end{figure*}

\begin{figure*}[!htbp]
    \centering
    \includegraphics[width=\textwidth]{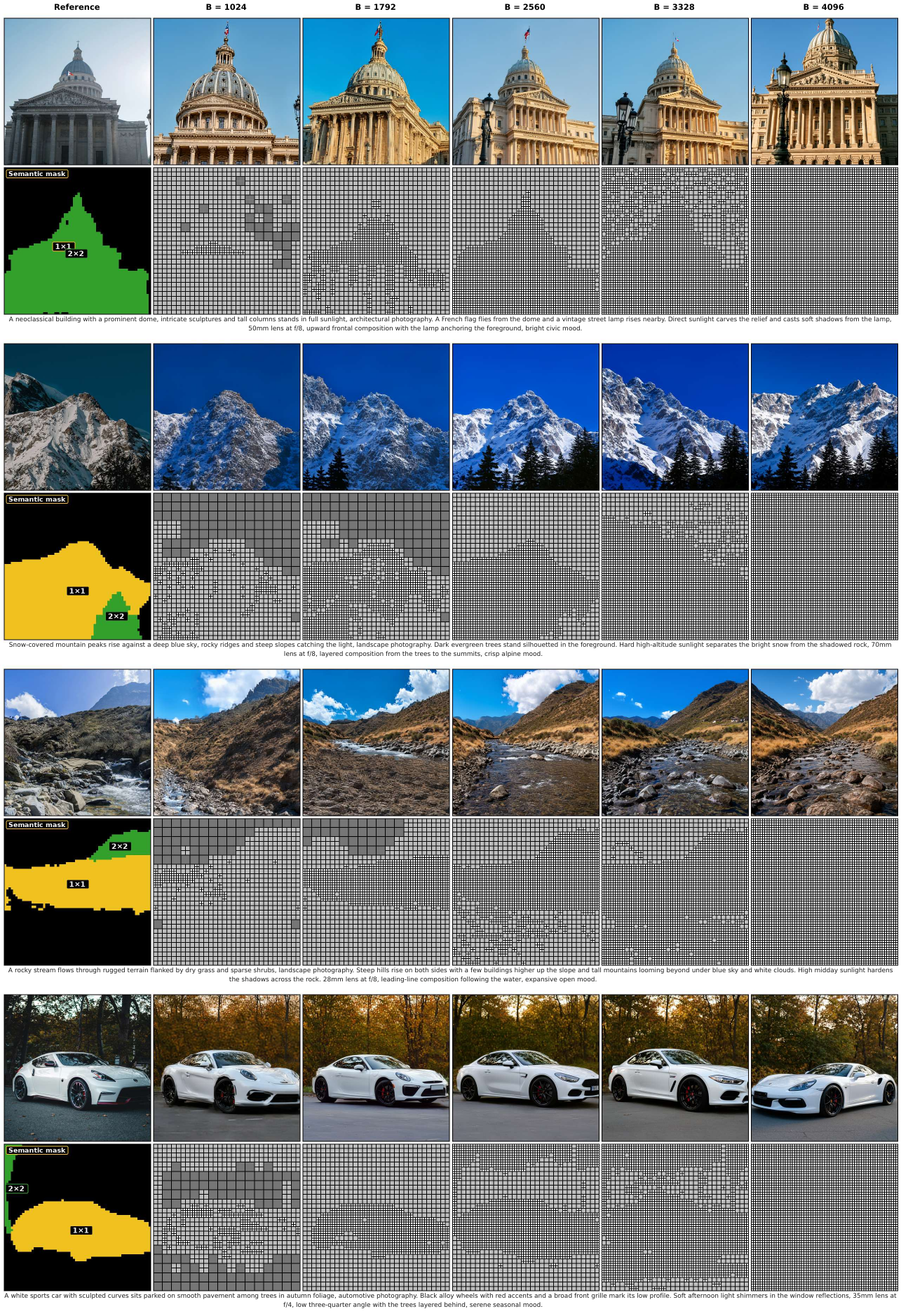}
    \caption{\textbf{Mask-guided Image Generation with Varying Token Budget.}}\label{fig:supp_budget_mask_2}
\end{figure*}

\begin{figure*}[!htbp]
    \centering
    \includegraphics[width=\textwidth]{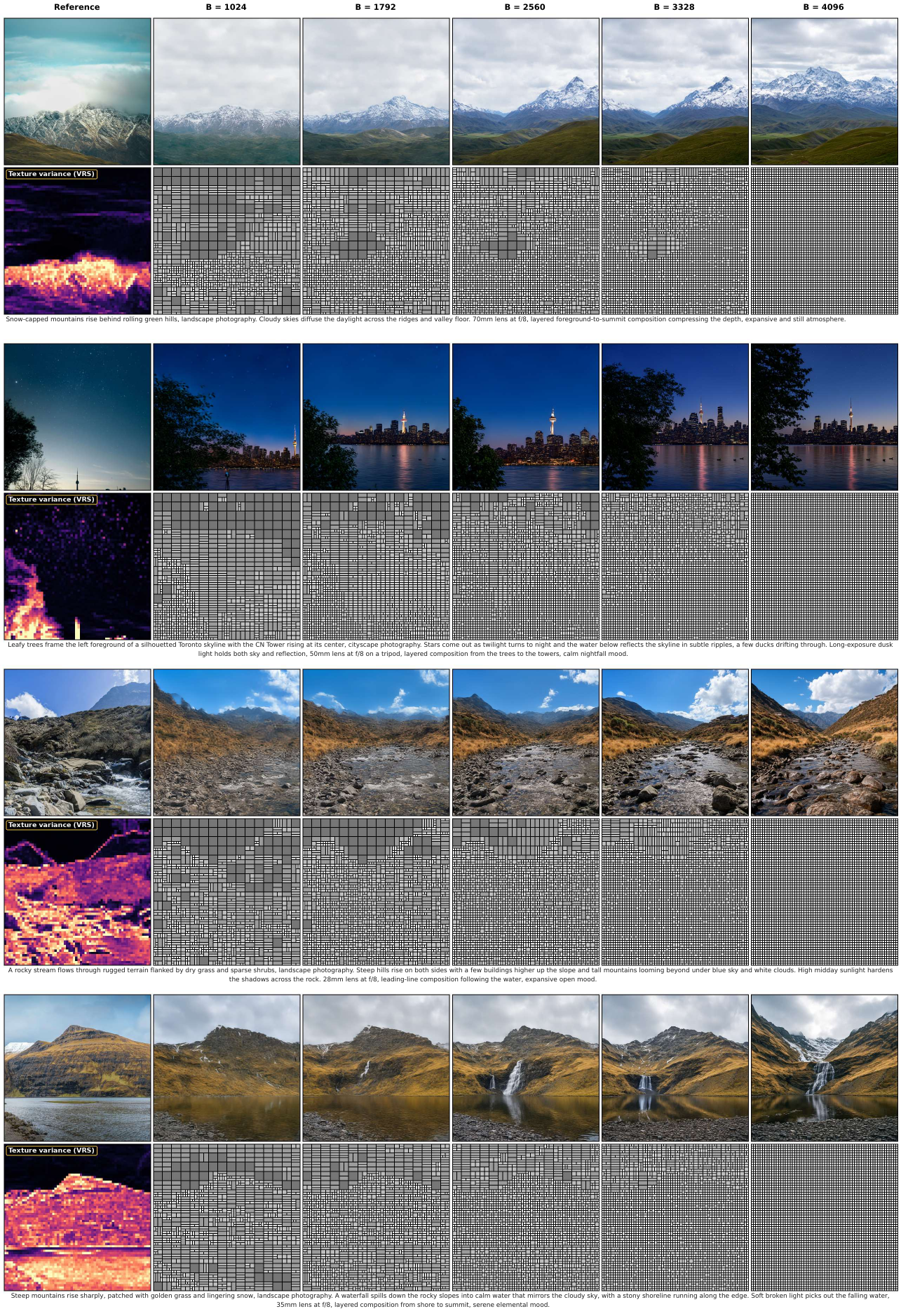}
    \caption{\textbf{VRS-guided Image Generation with Varying Token Budget -- Part 1.}}\label{fig:supp_budget_vrs_1}
\end{figure*}

\begin{figure*}[!htbp]
    \centering
    \includegraphics[width=\textwidth]{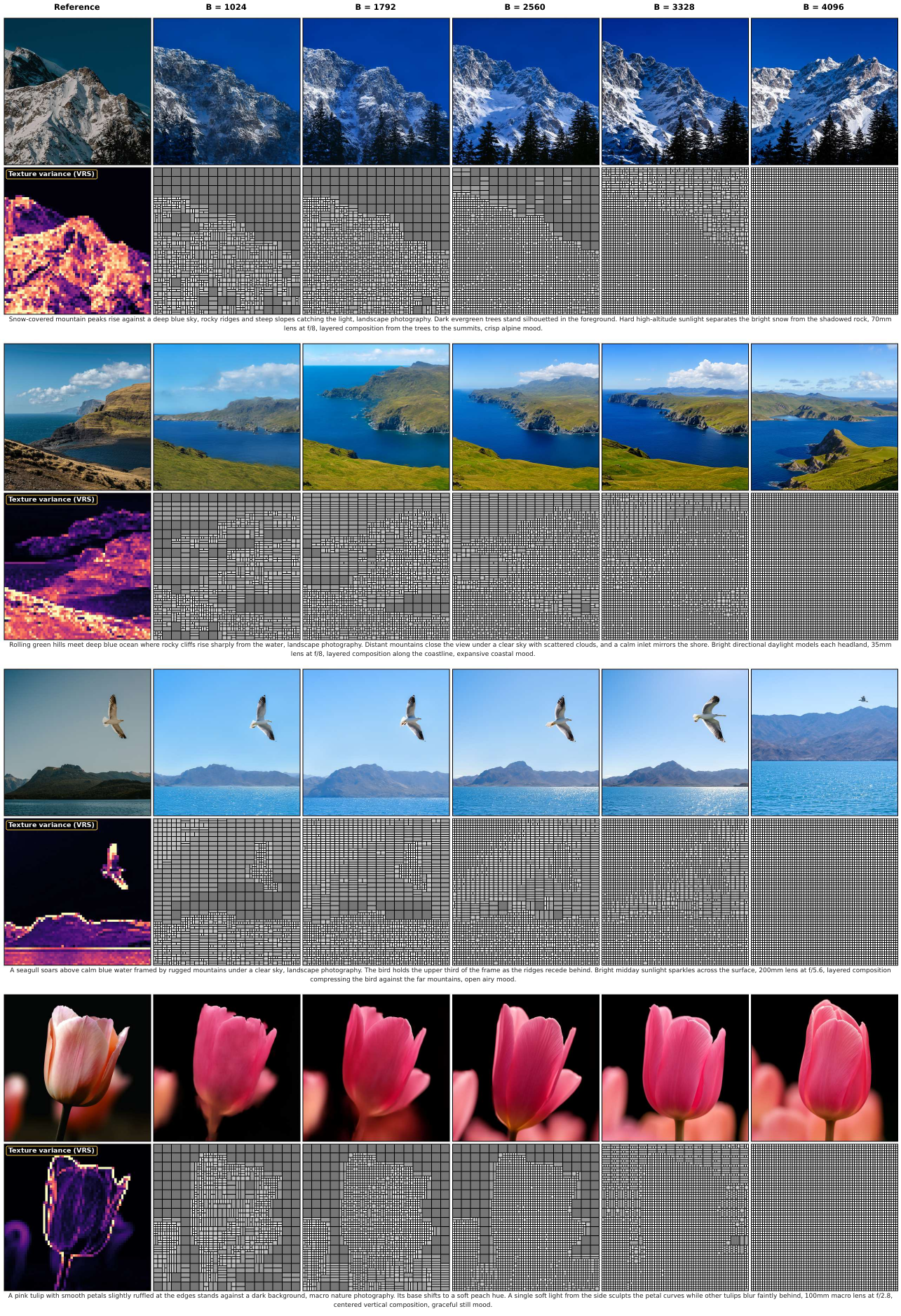}
    \caption{\textbf{VRS-guided Image Generation with Varying Token Budget -- Part 2}}\label{fig:supp_budget_vrs_2}
\end{figure*}

Figures~\ref{fig:supp_budget_mask_1}--\ref{fig:supp_budget_vrs_2} show how generation changes as the target token budget $B$ varies for the same prompt. We guide token allocation using semantic masks in Figures~\ref{fig:supp_budget_mask_1} and~\ref{fig:supp_budget_mask_2}, and texture variance in Figures~\ref{fig:supp_budget_vrs_1} and~\ref{fig:supp_budget_vrs_2}. Lower budgets use larger tokens over more of the image, retaining recognizable subjects and scene structure while simplifying fine textures. Higher budgets support finer spatial detail, such as foliage, fur, and architectural features. These results illustrate a controllable trade-off between computational budget and visual detail within the same model.

\FloatBarrier
\subsection{Additional Layout-Adaptive Image Generation Results}
\label{sec:supp:layout-image-results}
\begin{figure*}[!htbp]
    \centering
    \includegraphics[width=0.95\textwidth]{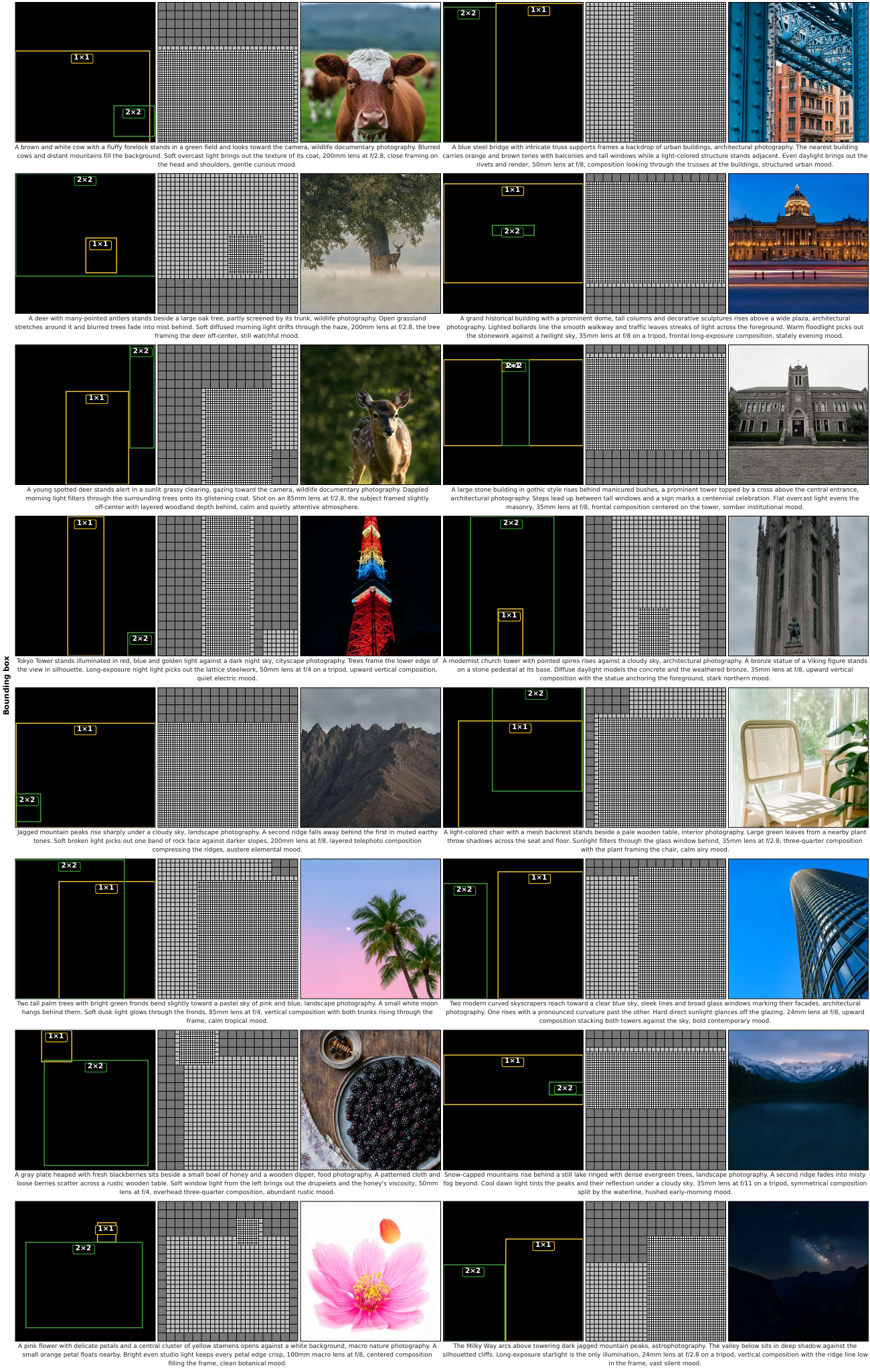}
    \caption{\textbf{Bounding-box–controlled Generation Results}}\label{fig:layout_supp_box}
\end{figure*}

\begin{figure*}[!htbp]
    \centering
    \includegraphics[width=0.95\textwidth]{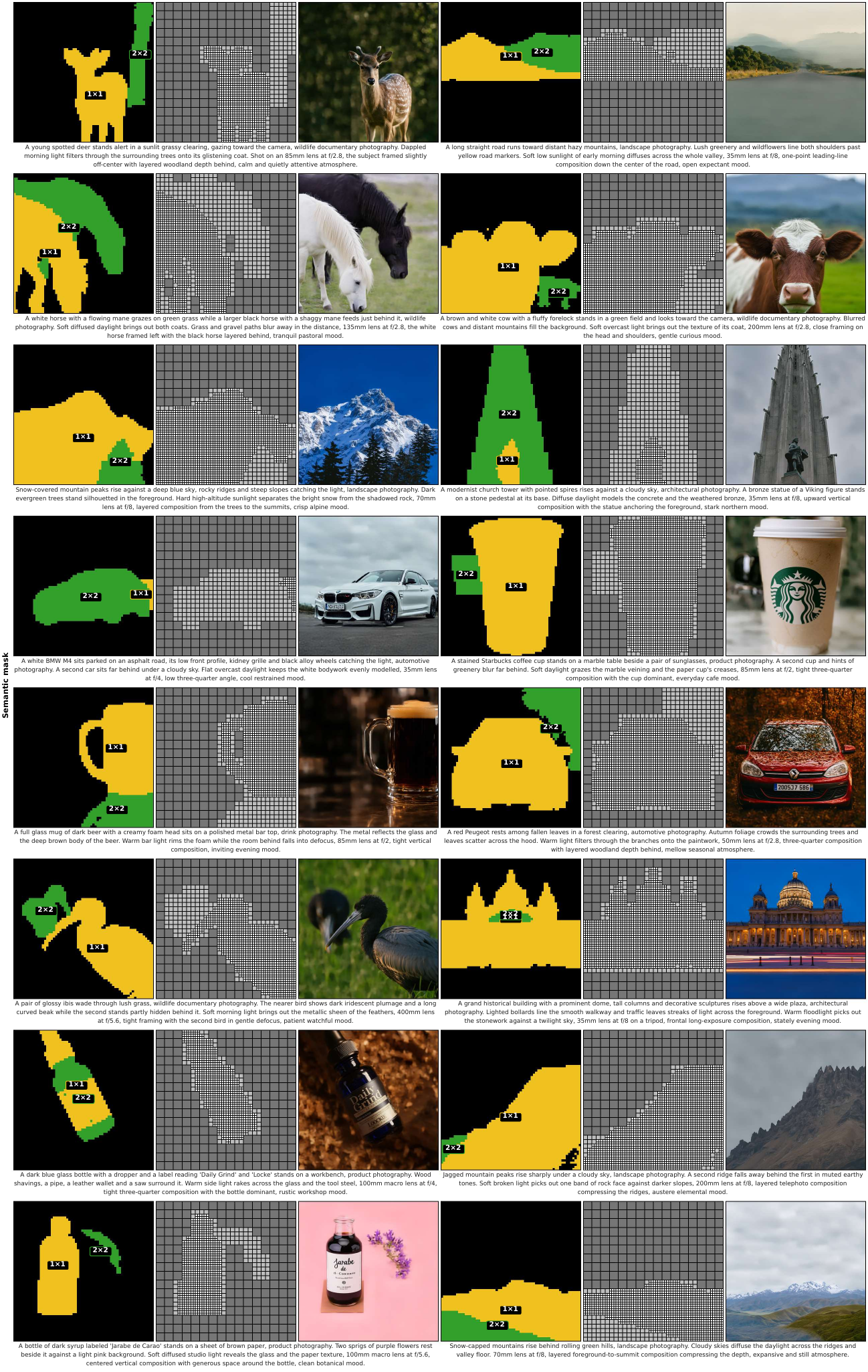}
    \caption{\textbf{Semantic-mask–controlled Generation Results}}\label{fig:layout_supp_mask}
\end{figure*}

\begin{figure*}[!htbp]
    \centering
    \includegraphics[width=0.95\textwidth]{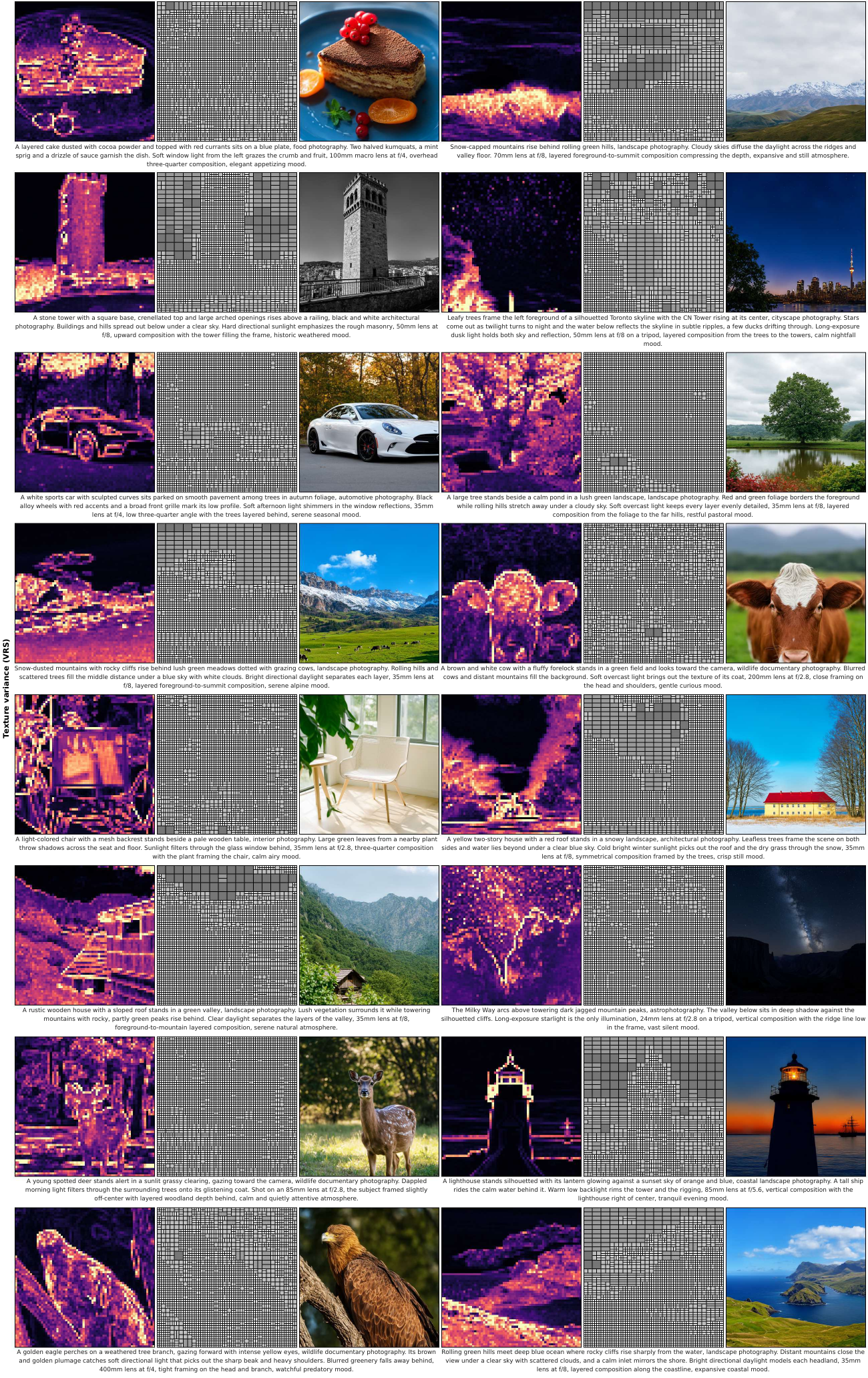}
    \caption{\textbf{Texture-variance–controlled Generation Results}}\label{fig:layout_supp_vrs}
\end{figure*}

\begin{figure*}[!htbp]
    \centering
    \includegraphics[width=0.9\textwidth]{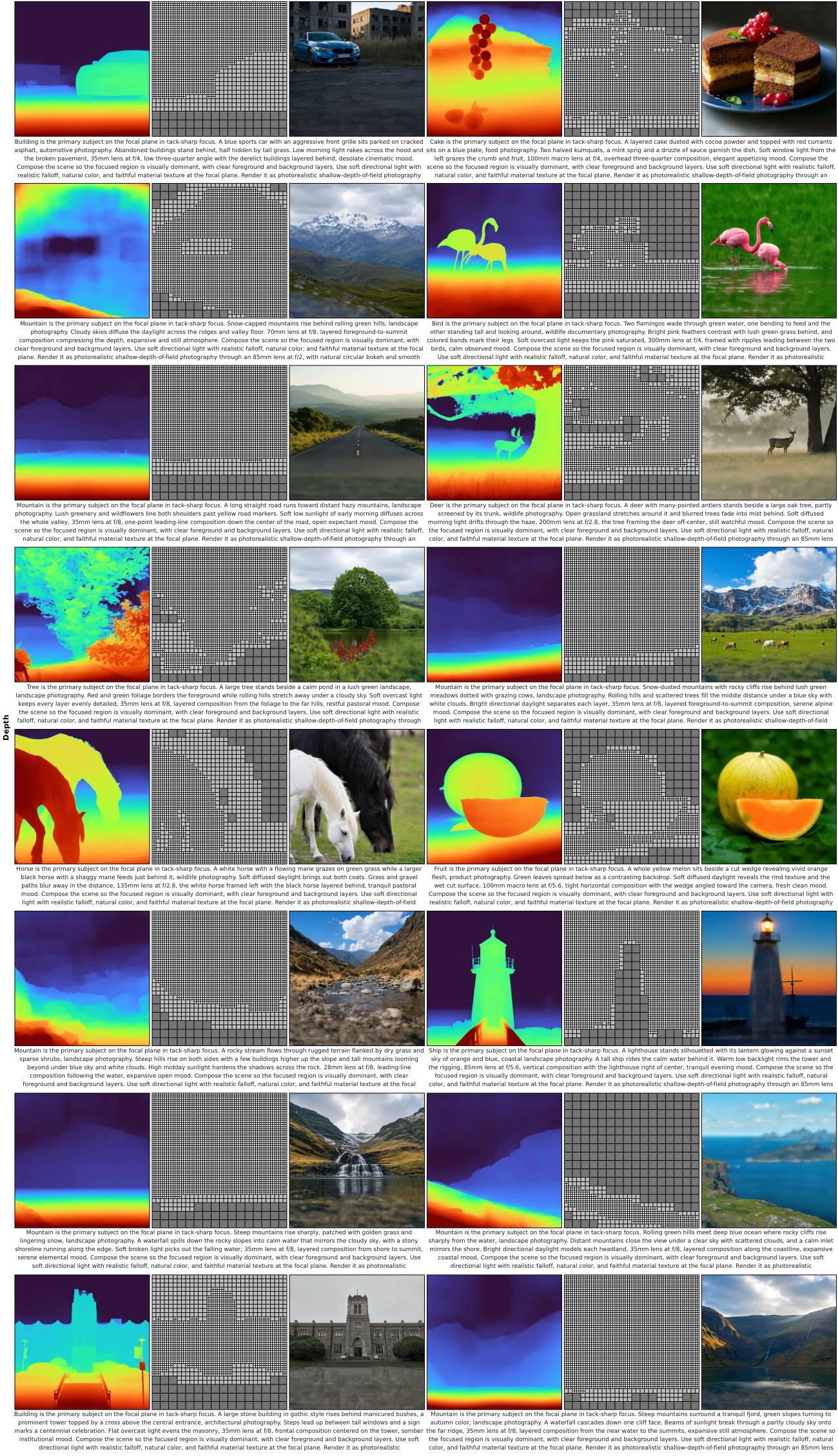}
    \caption{\textbf{Depth-of-field–controlled Generation Results}}\label{fig:layout_supp_dof}
\end{figure*}

Figures~\ref{fig:layout_supp_box}--\ref{fig:layout_supp_dof} show additional image generation examples generated by LoT-FLUX.2-4B using LoT layouts derived from different sources. Bounding boxes and semantic masks concentrate fine tokens in selected regions, where the model places salient subjects and detailed structures. Texture-variance layouts support spatially varying detail, while depth-derived layouts favor sharp content near the focal region and smoother, defocused surroundings. Additional layout-adaptive image generation results are available on the project website.

\FloatBarrier
\subsection{Layout Controllability Compared with Dense Generation}
\label{sec:supp:layout-controllability}
We evaluate whether changing the token layout changes where the generated
subject appears, comparing the mixed-source LoT-FLUX.2-4B and 9B models with
their corresponding pretrained full-resolution FLUX.2 models.

\paragraph{Experimental setup.}
We select 2,096 prompts from COCO-10k that describe one main subject and
have an unambiguous corresponding COCO bounding box. For each prompt, we
construct two layouts: one using the reference box and one randomly
translating it while preserving its size. We allocate $1\times1$ tokens
inside the box and $2\times2$ tokens outside; the paired layouts have
identical token counts. Both model sizes use the same prompts, layouts,
and seeds. LoT generates both images with the same prompt
and initial noise, changing only the layout. Dense FLUX.2 uses the same
prompt and initial noise but receives no layout; its single output is
evaluated against both target regions. This comparison measures the
additional spatial adaptability afforded by the token layout, rather
than comparing two layout-conditioned models.

\paragraph{Metrics and results.}
We use SAM3~\cite{sam3} with the subject name, without providing the target
box, and retain the highest-confidence subject mask above a confidence
threshold of $0.5$. \emph{Subject detection} is the percentage of images
with a detected subject. \emph{Target-region coverage} is the fraction of
the detected subject mask lying inside the target box,
$|M\cap R|/|M|$, with zero assigned when no subject is detected.
Tab.~\ref{tab:layout-controllability} shows higher coverage for LoT under
both layouts, with overall gains of 15.79 and 21.89 percentage points for
4B and 9B, respectively. Subject detection is slightly lower for LoT at
4B and comparable at 9B. Thus, subjects more often occupy the fine-token
region when generation can adapt to the supplied layout.

\begin{figure*}[p]
    \centering
    \includegraphics[width=\textwidth,height=0.90\textheight,keepaspectratio]{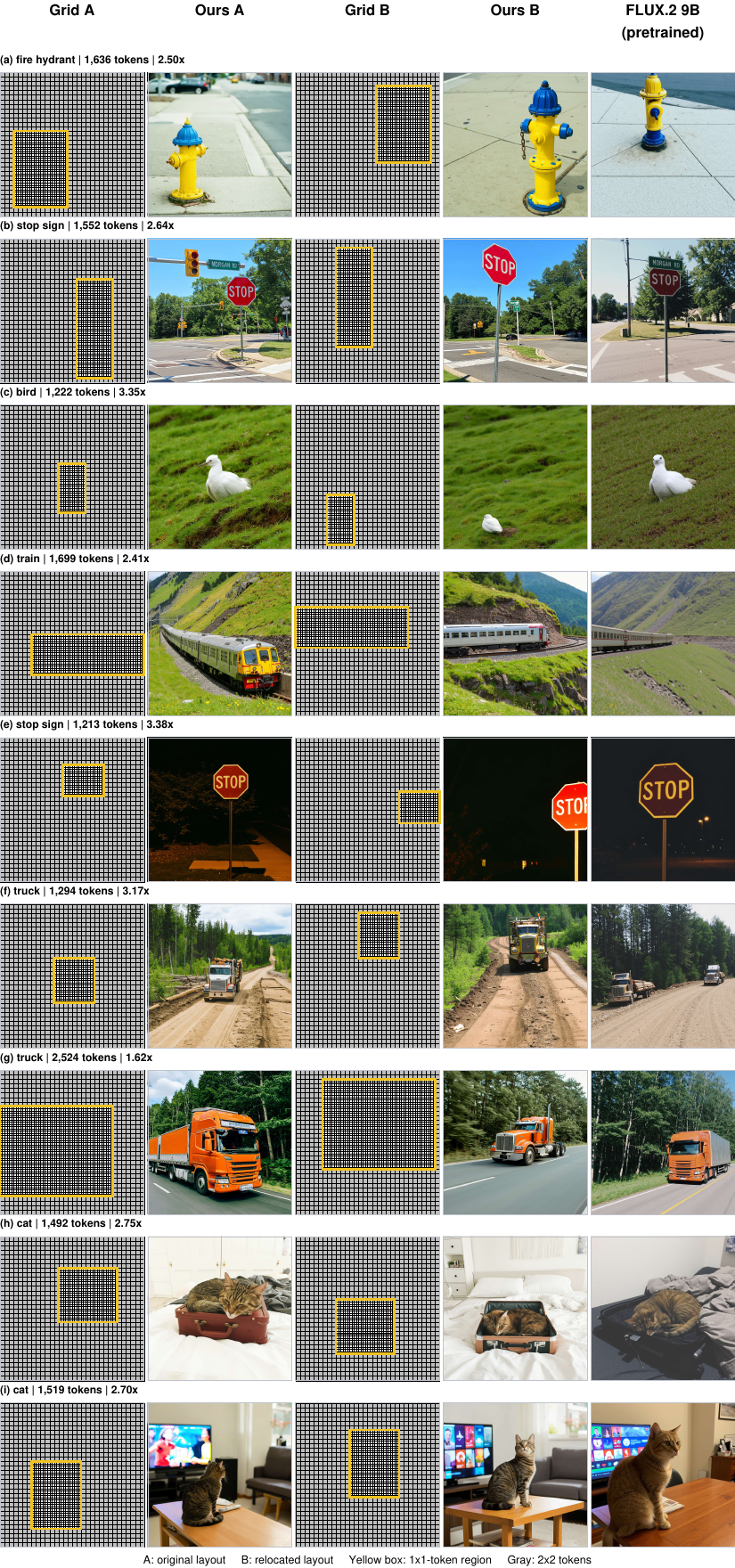}
    \caption{\textbf{Layout Controllability: LoT-FLUX.2 9B and Dense FLUX.2 9B.}
    Selected comparisons (1/2): original grid (A), ours (A),
    relocated grid (B), ours (B), and pretrained FLUX.2 9B. Each row shares the prompt
    and initial noise; A and B also share the token budget.
    Yellow boxes mark the $1\times1$-token regions in the layouts.
    Dense FLUX.2 receives no layout.}
    \label{fig:supp:layout-controllability}
\end{figure*}

\begin{figure*}[p]
    \centering
    \includegraphics[width=\textwidth,height=0.90\textheight,keepaspectratio]{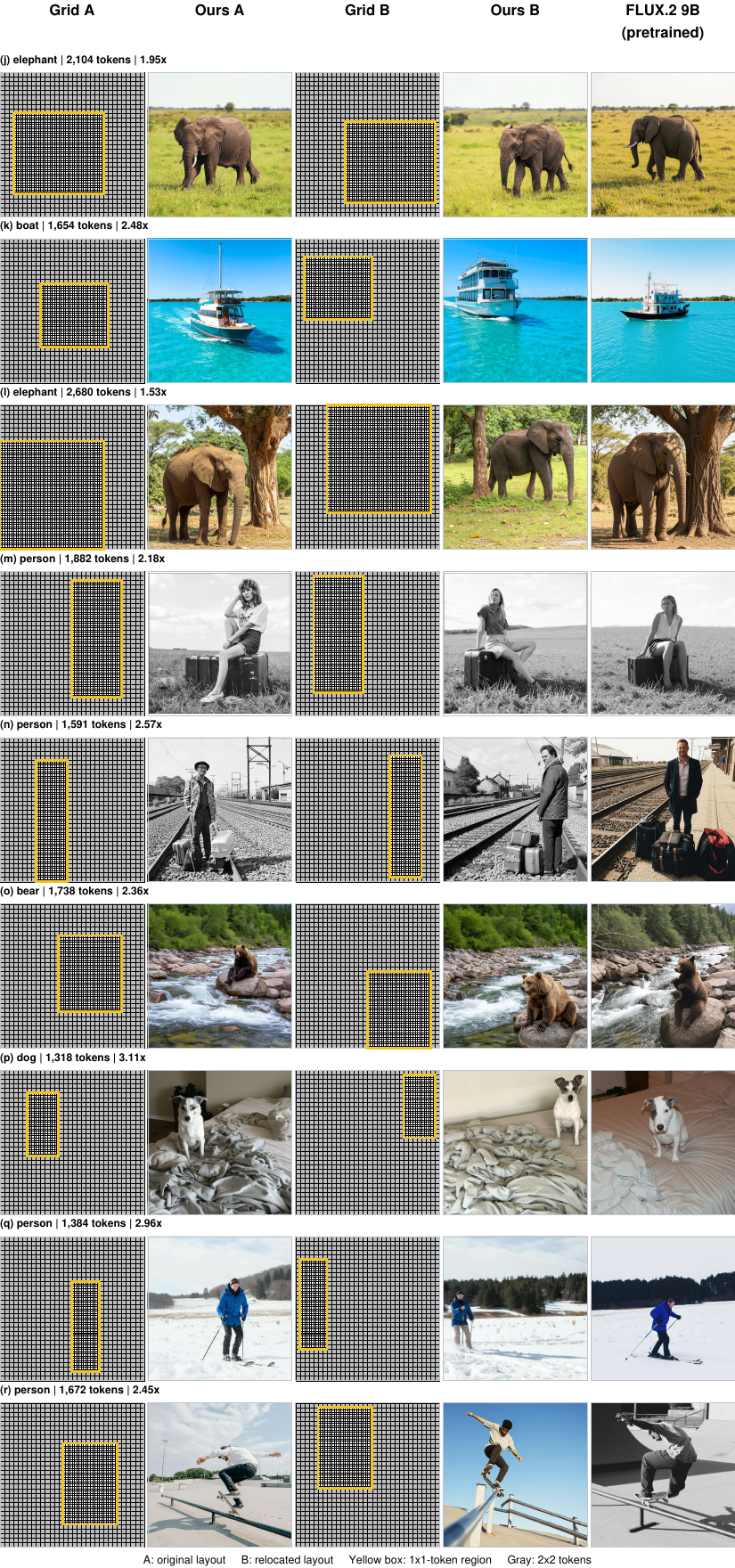}
    \caption{\textbf{Layout Controllability: LoT-FLUX.2 9B and Dense FLUX.2 9B.}
    Selected comparisons (2/2), following the same arrangement and protocol
    as Figure~\ref{fig:supp:layout-controllability}.}
    \label{fig:supp:layout-controllability-2}
\end{figure*}

Figures~\ref{fig:supp:layout-controllability} and~\ref{fig:supp:layout-controllability-2} provide selected qualitative
comparisons. With the prompt, initial noise, and token count fixed,
relocating the fine-token region changes the spatial placement of the
subject in LoT's outputs; the dense result is shown alongside for reference.

\begin{table}[!htbp]
    \centering
    \begin{tabular}{lrrrr}
        \toprule
         &  & \multicolumn{3}{c}{Target-region coverage} \\
        \cmidrule(lr){3-5}
        Method & Subject detection & Original & Relocated & Overall \\
        \midrule
        FLUX.2 4B (full res.) & \textbf{98.57} & 53.52 & 37.24 & 45.38 \\
        LoT-FLUX.2-4B & 98.12 & \textbf{68.84} & \textbf{53.50} & \textbf{61.17} \\
        \midrule
        FLUX.2 9B (full res.) & 98.95 & 52.98 & 36.90 & 44.94 \\
        LoT-FLUX.2 9B & \textbf{99.07} & \textbf{74.74} & \textbf{58.92} & \textbf{66.83} \\
        \bottomrule
    \end{tabular}
    \caption{\textbf{Layout Controllability Compared with Dense Generation.}
    All values are percentages (higher is better).}
    \label{tab:layout-controllability}
\end{table}

\FloatBarrier
\FloatBarrier
\FloatBarrier
\subsection{Video Generation Training Details}\label{sec:supp:video-training}
We fine-tune Wan2.1-T2V-14B~\cite{wan2025wan} with its native latent
normalization and frozen UMT5-XXL text encoder and VAE\@.

\paragraph{Training data and source mixture.}
We use 81-frame clips from Vchitect-T2V-Dataverse~\cite{fan2025vchitect}.
The layout mixture comprises semantic masks or bounding boxes (50\%, split
equally), VRS layouts (25\%), and DoF layouts (25\%). All training clips
use $1280\times720$ resolution.
Grids retain $e_t=1$ and $e_h,e_w\in\{1,2,4\}$, with at most 3,072 tokens
per latent time slice to avoid out-of-memory errors.

\paragraph{Training setup.}
We reuse the LoRA rank/dropout, AdamW coefficients, adapter learning rate
and weight decay, clamped clean-data loss, and EMA schedule
from~\Secref{sec:supp:training}. Wan-specific LoRA targets are the
query/key/value/output projections in self- and cross-attention, the
feed-forward linear layers, and the timestep embedding/projection layers.
We fully fine-tune the per-extent input embeddings, output heads, and
patch-size MLP at $10^{-5}$, and the final output normalization and
modulation at $10^{-4}$. These fully fine-tuned groups have zero weight decay.
We train for 4,000 steps with a total batch size of 32 videos,
using FSDP2, gradient checkpointing, and gradient-norm clipping
at 200. Unlike the image model's length-dependent dynamic shift, video
training uses a fixed noise-schedule shift factor of 5, shared by all tokens
within each clip.

\subsection{Video Generation Baseline Comparison Details}
\label{sec:supp:video-comparison-details}

\paragraph{Inference parameters.}
For all baselines, we generate with 50 sampling steps,
classifier-free guidance scale 4, orthogonal guidance strength 1, and a
noise-schedule shift factor of 5. We evaluate on 200 held-out DataVerse
test prompts, generating 81-frame videos at $1280\times720$ and 15 fps.
We use the test prompts to generate reference videos with
MiniMax-H3~\cite{minimax2026h3} and derive the evaluation VRS layouts from
these generated videos, rather than the original held-out clips. The
original clips were mostly static, so we use generated references to
evaluate temporally varying token allocations. Only the derived layouts,
not the reference RGB videos, are supplied to the evaluated models.
At each token budget, ours, Foveated Diffusion, and ToMe-SD use the same
VRS-derived grids, prompts, and seeds.

\paragraph{Training Details.}
We adapt Foveated Diffusion~\cite{chao2026foveateddiffusion} to Wan2.1 14B
and the same VRS-derived token layouts as ours, with spatial extents up to
$4\times4$. We evaluate our VRS-only model and Foveated Diffusion at their
1,000-step checkpoints.
ToMe-SD~\cite{bolya2023token} requires no training: we adapt token merging
to pretrained Wan2.1 14B by mean-merging tokens within each VRS leaf for
self-attention, cross-attention, and MLP computation, then copy-unmerging
the outputs before adding them to the full-resolution residual stream.
Reported Full Res. Wan2.1 14B is fine-tuned using the same training budget and parameters on the same dataset for fairness. 

\paragraph{Metric details.}
We evaluate six VBench~\cite{huang2023vbench} dimensions using its
custom-video evaluation on our held-out prompts.
For image quality, aesthetic quality measures visual appeal, while imaging
quality assesses perceptual degradation such as blur and noise.
For temporal consistency, subject consistency measures the stability of
subject appearance across frames, while background consistency measures
the stability of the scene background.
For motion, motion smoothness measures the continuity of movement, while
dynamic degree measures the prevalence of substantial motion.
Higher quality, consistency, and smoothness scores are better; a higher
dynamic degree indicates more motion, not necessarily more realistic motion.

For efficiency, runtime is the mean H100 generation time per video that includes both denoising and VAE decoding, similar to the image generation case. 

\subsection{Additional Video Results}
\label{sec:supp:video-results}
The agent-assisted video demos on the project website demonstrate custom video layout control: users paint a static background detail map and keyframed detail maps to specify where fine and coarse tokens are allocated as subjects move.

Additional qualitative video generation results on the project website cover
semantic masks, bounding boxes, texture variance, and depth-of-field layouts.
Each example pairs its source cue and token layout with the generated video.
The video baseline comparison setup is reported in
Sec.~\ref{sec:supp:video-comparison-details}.

\subsection{Additional Ablation Study Results}
\label{sec:supp:ablation}
Tab.~\ref{tab:ablation_dit_supp} reports ablation results across different
token budgets, extending the main-paper results in Tab.~\ref{tab:ablation_dit},
which reports only a single threshold setting at approximately
$1.52\times$ token compression.
We use the same training and evaluation setup as the image generation
baseline comparison. We introduce the ablation variants considered below.

\paragraph{Ours w/ direct HR prediction.}
We bypass the analytic patch-wise asymmetric flow recovery
in~\eqref{eq:method-asym-recovery}, introduced in~\Secref{sec:method:asymflow}, and directly
interpret the per-extent output heads as the full-resolution flow
prediction. The compressed input representation, transformer, and
patch-size conditioning are retained. The heads are trained against the
full-resolution flow objective without the analytic recovery step.

\paragraph{Ours w/ binary grid.}
We restrict the layout to square $1\times1$ and $2\times2$ tokens.
This replaces the multilevel, potentially
anisotropic layout during both training and evaluation; patch-size
conditioning and asymmetric flow recovery remain unchanged.

\paragraph{Ours w/o size embedding.}
We disable the learned patch-size embedding introduced
in~\Secref{sec:method:dit}, removing explicit
extent-dependent conditioning of the transformer. We retain the
extent-specific input and output heads, single-scale RoPE at each token's
geometric center, and asymmetric flow recovery. This isolates the effect
of explicitly informing the transformer of each token's spatial extent.

\begin{figure*}[!htbp]
    \centering
    \includegraphics[width=0.95\textwidth]{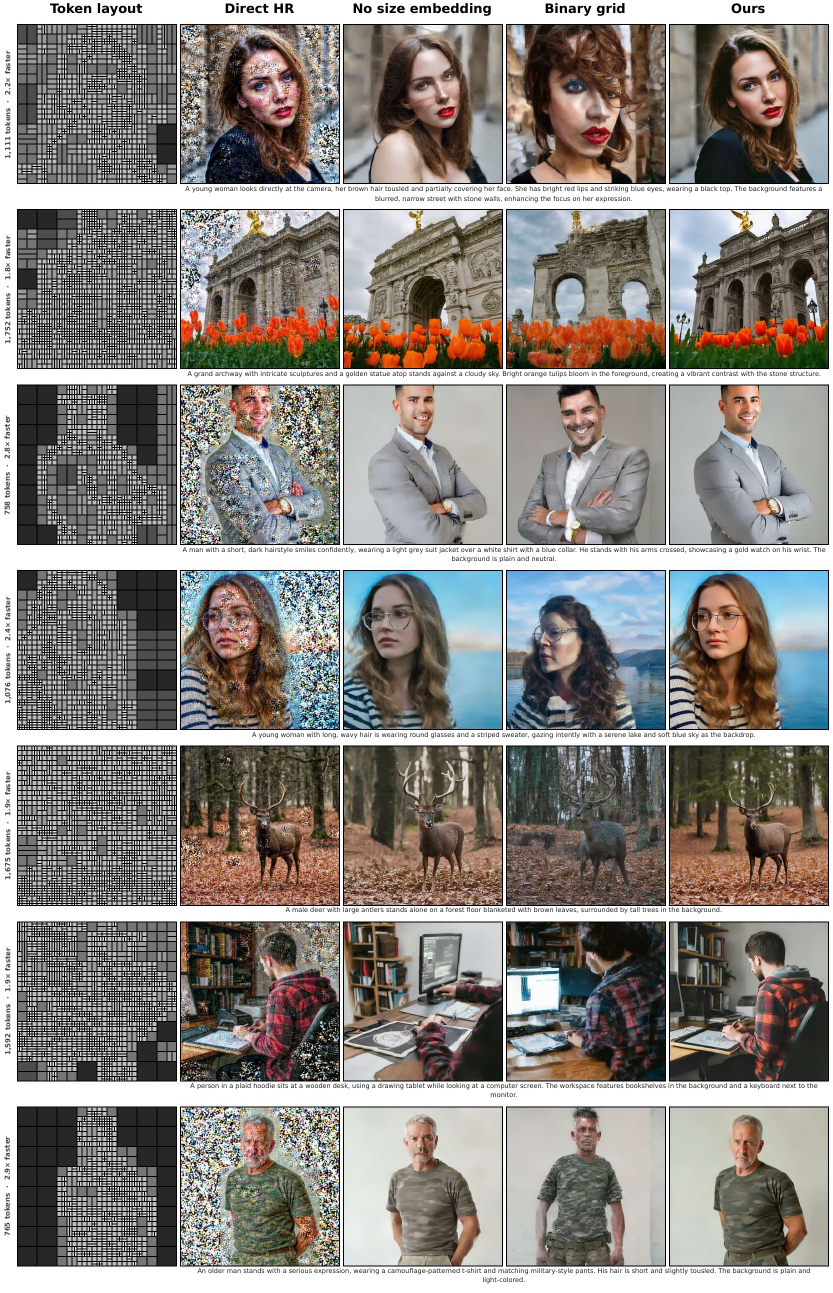}
    \caption{\textbf{Ablation Study Qualitative Results.}}\label{fig:ablation_qual}
\end{figure*}

We show image generation qualitative results of the architectural ablation studies in Fig. \ref{fig:ablation_qual}.

\begin{table*}[t]
    \centering
    \resizebox{\textwidth}{!}{%
    \begin{tabular}{lcccccccccc}
        \toprule
        Method & Compression & Time (s) $\downarrow$ & Speed $\uparrow$ & IR $\uparrow$ & HPSv2.1 $\uparrow$ & HPSv3 $\uparrow$ & FID $\downarrow$ & pFID $\downarrow$ & TOPIQ $\uparrow$ & MUSIQ $\uparrow$ \\
        \midrule
        Ours w/o size embedding & 1.523$\times$ & 4.372 & 1.322$\times$ & \textbf{0.9294} & \underline{0.2717} & \underline{10.0284} & \underline{14.584} & \underline{17.472} & \underline{0.6030} & \underline{69.433} \\
        Ours w/ direct HR prediction & 1.523$\times$ & \underline{4.264} & \underline{1.355$\times$} & 0.6301 & 0.2601 & 7.0595 & 26.487 & 36.072 & 0.5134 & 63.346 \\
        Ours w/ binary grid & 1.518$\times$ & \textbf{4.167} & \textbf{1.387$\times$} & 0.7472 & 0.2520 & 8.1955 & 20.603 & 25.073 & 0.5517 & 66.443 \\
        Ours & 1.523$\times$ & 4.366 & 1.323$\times$ & \underline{0.9105} & \textbf{0.2764} & \textbf{10.3477} & \textbf{13.800} & \textbf{17.019} & \textbf{0.6170} & \textbf{70.410} \\
        \midrule
        Ours w/o size embedding & 1.893$\times$ & 3.718 & 1.554$\times$ & \textbf{0.9008} & \underline{0.2665} & \underline{9.7299} & \underline{13.934} & \underline{17.320} & \underline{0.5888} & \underline{68.563} \\
        Ours w/ direct HR prediction & 1.893$\times$ & \underline{3.609} & \underline{1.601$\times$} & 0.4335 & 0.2503 & 5.3532 & 37.414 & 55.443 & 0.4581 & 59.576 \\
        Ours w/ binary grid & 1.879$\times$ & \textbf{3.501} & \textbf{1.651$\times$} & 0.7086 & 0.2478 & 7.8377 & 21.638 & 27.603 & 0.5180 & 64.668 \\
        Ours & 1.893$\times$ & 3.704 & 1.560$\times$ & \underline{0.8833} & \textbf{0.2728} & \textbf{10.1211} & \textbf{13.361} & \textbf{16.393} & \textbf{0.6080} & \textbf{69.691} \\
        \midrule
        Ours w/o size embedding & 2.406$\times$ & 3.137 & 1.842$\times$ & \underline{0.8337} & \underline{0.2590} & \underline{9.1663} & \underline{14.378} & \underline{18.822} & \underline{0.5758} & \underline{67.968} \\
        Ours w/ direct HR prediction & 2.406$\times$ & \underline{3.071} & \underline{1.882$\times$} & 0.1066 & 0.2351 & 2.8754 & 54.762 & 89.686 & 0.3876 & 54.603 \\
        Ours w/ binary grid & 2.336$\times$ & \textbf{2.958} & \textbf{1.953$\times$} & 0.6952 & 0.2464 & 7.6611 & 23.111 & 31.550 & 0.4871 & 63.340 \\
        Ours & 2.406$\times$ & 3.178 & 1.818$\times$ & \textbf{0.8513} & \textbf{0.2672} & \textbf{9.7533} & \textbf{13.308} & \textbf{17.046} & \textbf{0.5978} & \textbf{69.420} \\
        \midrule
        Ours w/o size embedding & 2.935$\times$ & 2.789 & 2.071$\times$ & \underline{0.8042} & \underline{0.2543} & \underline{8.7670} & \underline{14.933} & \underline{20.497} & \underline{0.5636} & \underline{67.277} \\
        Ours w/ direct HR prediction & 2.935$\times$ & \underline{2.704} & \underline{2.137$\times$} & -0.0843 & 0.2268 & 1.3967 & 68.508 & 111.952 & 0.3497 & 51.809 \\
        Ours w/ binary grid & 2.774$\times$ & \textbf{2.634} & \textbf{2.194$\times$} & 0.6793 & 0.2460 & 7.5753 & 24.004 & 34.786 & 0.4674 & 62.508 \\
        Ours & 2.935$\times$ & 2.832 & 2.040$\times$ & \textbf{0.8326} & \textbf{0.2639} & \textbf{9.5522} & \textbf{13.345} & \textbf{17.822} & \textbf{0.5873} & \textbf{68.760} \\
        \bottomrule
    \end{tabular}%
    }
    \caption{\textbf{Ablation Study Across Different Token Budgets.}}\label{tab:ablation_dit_supp}
\end{table*}

\FloatBarrier
\subsection{Limitations}
\label{sec:supp:limitation}

\begin{figure*}[!htbp]
    \centering
    \includegraphics[width=\textwidth]{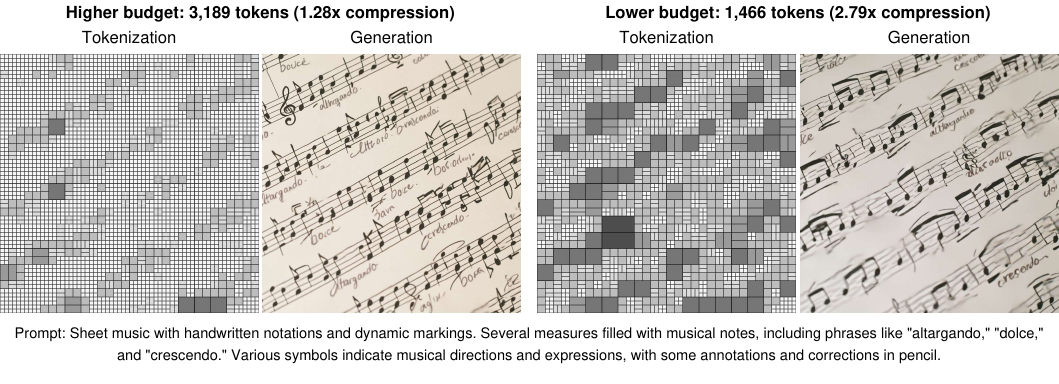}
    \caption{\textbf{Limitation: Fine-Structure Degradation at Lower Token Budgets.}
    The lower-budget result shows more broken
    staff lines.}
    \label{fig:limitation_music}
\end{figure*}

LoT gains efficiency by allocating fewer tokens to regions that can be synthesized more coarsely. When fine structures or textures extend across most of an image, as in dense musical notation or close-up views of grass or fabric, preserving them may require a fine token layout throughout, leaving limited room for computational savings. Enforcing a lower token budget in these cases can degrade structural fidelity. As shown in Fig.~\ref{fig:limitation_music}, the lower-budget result exhibits more fragmented and distorted staff lines than the higher-budget result generated with the same prompt and noise seed. Thus, the achievable savings depend on the scene's detail requirements, and layout adaptation does not guarantee faithful fine structures under aggressive token reduction.

\end{document}